\documentclass[a4paper,fleqn]{cas-sc}
\usepackage{etoolbox}
\makeatletter
\patchcmd{\ps@pprintTitle}{\footnotesize\itshape
       Preprint submitted to \ifx\@journal\@empty Elsevier
       \else\@journal\fi\hfill\today}{\relax}{}{}
\makeatother
\usepackage{color}
\usepackage{longtable}
\usepackage{xcolor}
\usepackage{blindtext}

\usepackage{amssymb,bezier,amsfonts, enumerate, keyval}
\usepackage{amssymb,bezier,times} 
\usepackage{color}
\usepackage[shortlabels]{enumitem}
\usepackage[ruled,norelsize]{algorithm2e}
\usepackage{multirow}
\usepackage{float}
\usepackage[caption = false]{subfig}
\usepackage{amsthm}
\usepackage{soul}

\theoremstyle{definition}

\makeatletter
\newcommand{\removelatexerror}{\let\@latex@error\@gobble}
\makeatother

\makeatletter
\let\oldlt\longtable
\let\endoldlt\endlongtable
\def\longtable{\@ifnextchar[\longtable@i \longtable@ii}
\def\longtable@i[#1]{\begin{figure}[t]
\onecolumn
\begin{minipage}{0.5\textwidth}
\oldlt[#1]
}
\def\longtable@ii{\begin{figure}[t]
\onecolumn
\begin{minipage}{0.5\textwidth}
\oldlt
}
\def\endlongtable{\endoldlt
\end{minipage}
\twocolumn
\end{figure}}
\makeatother

\usepackage[numbers]{natbib}

\def\tsc#1{\csdef{#1}{\textsc{\lowercase{#1}}\xspace}}
\tsc{WGM}
\tsc{QE}
\tsc{EP}
\tsc{PMS}
\tsc{BEC}
\tsc{DE}

\begin{document}
\let\WriteBookmarks\relax
\def\floatpagepagefraction{1}
\def\textpagefraction{.001}
\shorttitle{Reliability Check via Weight Similarity in Privacy-Preserving Multi-Party Machine Learning}
\shortauthors{Edemacu et~al.}

\title [mode = title]{Reliability Check via Weight Similarity in Privacy-Preserving Multi-Party Machine Learning}
                     

%
%

\author{Kennedy~Edemacu}
\ead{k.edemacu@muni.ac.ug}




\author{Beakcheol~Jang}
\cormark[1]
\ead{bjang@smu.ac.kr}



\author{Jong~Wook~Kim}
\cormark[1]
\ead{jkim@smu.ac.kr}

\address{Department of Computer Science, Sangmyung University, Seoul (South Korea).}

\cortext[cor1]{Corresponding author}
%

\begin{abstract}
Multi-party machine learning is a paradigm in which multiple participants collaboratively train a machine learning model to achieve a common learning objective without sharing their privately owned data. The paradigm has recently received a lot of attention from the research community aimed at addressing its associated privacy concerns. In this work, we focus on addressing the concerns of data privacy, model privacy, and data quality associated with privacy-preserving multi-party machine learning, i.e., we present a scheme for privacy-preserving collaborative learning that checks the participants' data quality while guaranteeing data and model privacy. In particular, we propose a novel metric called \textit{weight similarity} that is securely computed and used to check whether a participant can be categorized as a reliable participant (holds good quality data) or not. The problems of model and data privacy are tackled by integrating homomorphic encryption in our scheme and uploading encrypted weights, which prevent leakages to the server and malicious participants, respectively. The analytical and experimental evaluations of our scheme demonstrate that it is accurate and ensures data and model privacy.
\end{abstract}

%
%

\maketitle

\section{Introduction}\label{sec:introduction}

Recently, machine learning techniques have achieved tremendous success in many applications (e.g., cancer detection \cite{cruz2017, kharya2016}, face recognition \cite{cao2018, deng2019}, speech recognition \cite{lim2016, agarwalla2016}, playing the game of Go \cite{silver2017}, etc.), with their performances equaling or surpassing that of humans. Central to this success is the demand for large amounts of data used in the training of the models. However, because of legal issues, competitive advantage and privacy concerns, collecting data from multiple sources is a challenging task. For one, centrally gathered data might be permanently stored and used without the data owner's knowledge, which is contrary for example to a regulation by EU \cite{gdpr} that grants data owners the right to ask companies to permanently delete their data. This results in data holding institutions' reluctance to share their data. Hence, a great hindrance to the success of machine learning as models get trained on limited amounts of data.

The recent advances of multi-party machine learning \cite{shokri2015, phong2018, aono2016, gong2020, zhao2019, phuong2019} have shown a promising potential in addressing the data scarcity challenge. In multi-party machine learning, multiple participants collaboratively train a common model while keeping their data private. The participants share the same model architecture and a common learning objective. The general training convention is that, first, global parameters are initialized and stored by a server. Each participant then downloads the global parameters for training its local model based on its private data and uploads the intermediate gradients or weights to the server for updating the global parameters. This is repeated until the common learning objective is achieved. Much as the paradigm prevents an attacker from having a direct access to participants' data, there are still several ways it can be compromised to:
\begin{enumerate}
  \item Reveal a participant's training data through, e.g., using a generative adversarial network by an attacker to deceive the participant into revealing detailed information during the training process \cite{hitaj2017}, uploaded gradients during the collaborative training process \cite{phong2018}, reverse-engineering the participant's local model \cite{pathak2010}, etc.
  \item Reveal the trained model, which can lead to the recovery of information about the training data through model inversion attacks \cite{zhang2020, fredrikson2015}, and
  \item Reduce the effectiveness and the training efficiency through injection of poor quality data during the training process.
\end{enumerate}

In literature, the training data and model leakage problems are mainly addressed through differential privacy and homomorphic encryption. Specifically, with differential privacy, leakage is prevented by adding noise to the data item. A parameter $\epsilon $ referred to as the privacy budget is used to control the amount of the added noise and the achieved privacy level. In \cite{shokri2015}, a differential technique is used to add noise to the uploaded gradients during collaborative learning. To prevent privacy leakages due to high privacy budget consumption, Gong \textit{et al.} \cite{gong2020} proposed a dynamic allocation of privacy budgets during multi-party learning. With homomorphic encryption, parameters are encrypted, and the algebraic operations for parameter updates are performed in an encrypted form. Additive homomorphic encryption algorithms are used for preventing information leakage to the central server in \cite{phong2018, aono2016}.
Besides differential privacy and homomorphic encryption, Zhang \textit{et al.} \cite{zhang2017} utilized the threshold secret sharing technique during global parameter updates in collaborative learning, i.e., the global parameter updates are only effected when the number of uploaded intermediate parameters reach a certain threshold. In \cite{phuong2019}, a symmetric encryption technique is used to prevent information leakage to the central server during multi-party training. Here, the central server simply acts as a relay point to convey parameters to the next participant to continue with the training process using its local dataset. The same work proposed the upload of intermediate weights instead of gradients. They proved that unlike gradients, weights reveal no information about the training data. 

The performance challenge attributed to poor data quality in privacy-preserving collaborative learning has remained an open problem, and to the best of our knowledge, the only work that has attempted to address it was conducted by Zhao \textit{et al.} in \cite{zhao2019}. In their work, a participant is categorized as a \textit{reliable participant} (RP) or as an \textit{unreliable participant} (UP). RPs hold good quality and similar data while UPs, which are assumed to be few in number, hold poor quality data. During the learning process, intermediate parameters are uploaded by the participants, and the server generates a utility score for each participant using a common validation dataset. The utility score shows the accuracy of each participant's parameters, and it is used to determine the participants whose parameters are included during the global parameter update. Although this approach is demonstrated to be effective in minimizing the influence of UPs, it has some limitations. First, the semi-honest server used in the work knows the validation dataset that is utilized to compute the utility scores for the participants and that can provide a clue about the general training dataset. Second, the semi-honest server is not prevented from accessing the trained model, which can lead to the recovery of information regarding the training data used to train the model. In a public setting, e.g., a cloud setting, these exposures can be detrimental. Thus, fixing these limitations through a novel scheme is a necessity.

Motivated by the preceding observations, we design a privacy-preserving multi-party machine learning scheme with the following features: (1) the proposed scheme leaks no private information to an honest-but-curious server or any participant (RP or UP) involved in the learning process. (2) the proposed scheme leaks no information about the trained model to the honest-but-curious server. And, (3) the proposed scheme minimizes the disruptions caused by UPs during the collaborative learning process. To defend against the honest-but-curious server and the participants while minimizing the effects of poor quality data from UPs, we utilize two main techniques. First, the additively homomorphic Paillier algorithm \cite{paillier1999}, which allows algebraic addition and subtraction operations to be correctly performed on ciphertexts. This enables the server to update the global parameters using the ciphertexts uploaded by the participants. Next, we propose a metric called the \textit{weight similarity} (more on this in section \ref{sec:weight}). In order to compute the weight similarity scores, we introduce an additional entity in our scheme called the \textit{model initiator}. The \textit{model initiator} is simply an RP who initiates the collaborative learning process. Similarity scores are computed between the \textit{model initiator}'s parameters and the other participants' parameters. Depending on whether the score is above a threshold, a participant's parameters might be included or excluded during the global parameter updates. This way, the disruptive influence of the UPs can be minimized during multi-party machine learning. Finally, to defend against information leakage to malicious participants, each participant uploads encrypted intermediate weights instead of gradients in the proposed scheme. Weights leak no information as proved by Phong and Phuong proved in \cite{phuong2019}. We summarize our contributions as follows.
\begin{itemize}
  \item To the best of our knowledge, we are the first to investigate the similarities between intermediate weights uploaded by participants in multi-party machine learning.
  \item We design a novel privacy-preserving multi-party machine learning scheme that integrates homomorphic encryption and weight similarity scores to prevent leakages to the central server and the participants while minimizing the disruptive influence of UPs during the collaborative learning process.
  \item We evaluate the performance of our proposed scheme on real-world datasets. The results demonstrate that our proposed scheme guarantees privacy and achieves high accuracy while being robust to UPs. 
\end{itemize}

The rest of the paper is organized as follows: in section \ref{sec:related_work}, we present the other related works. Section \ref{sec:preliminaries} discusses the preliminary concepts used in the work. In section \ref{sec:our_work}, we present our proposed multi-party machine learning system with reliability check. The section also contains the discussion of our proposed weight similarity metric. The experiments are presented in section \ref{sec:experiments} and section \ref{sec:conclusion} concludes the work.

\section{Other Related Works}\label{sec:related_work}
In \cite{gilad2016}, Gilad-Bachrach \textit{et al.} proposed a system referred to as CryptoNets for performing predictions on encrypted data. To make a prediction on a data item, the homomorphically encrypted data item is fed to a model that is already trained. The fed data goes through the feed-forward process of machine learning and the prediction results are returned in an encrypted form. In \cite{ma2019}, Ma \textit{et al.} presented a scheme for predictions on encrypted data using non-interactive neural networks. In their scheme, an already-trained model is split into two parts and each part is given to a server. To perform a prediction on an encrypted data item, the data item is also split into two parts and each server receives a part. The two servers interact to generate a prediction for the fed data item. Our work differs from \cite{gilad2016, ma2019} in the sense that, in our work, we aim at securely and accurately training the weights that can be used to perform predictions on data items, which is not the case in \cite{gilad2016, ma2019} in which the weights are already trained and are simply used to perform predictions on encrypted data items. 

Cao \textit{et al.} \cite{cao2020} presented a scheme for synchronous and parallel privacy-preserving collaborative learning in which the participants only send their local cost values to the server during global parameter updates. The server utilizes the received cost values to identify the participant with the best local model at a training round and requests its parameters. The server then updates the global parameters using the requested parameters. The authors claim their scheme prevents information leakage. However, their work does not exploit the full benefits of collaborative learning, since it only depends on the best performer at each training round. 

An integration of homomorphic encryption with proxy re-encryption for privacy-preserving multi-party learning is presented in \cite{zhang2019}. In this work, each participant has a unique key, and every parameter encrypted by the participants during collaborative learning has to be transformed using a proxy key for updating the global parameters. Their system requires an additional server and incurs high communication overhead. Aspects of data quality are not considered in this work as well. 

Bonawitz \textit{et al.} \cite{bonawitz2017} presented a scheme that securely aggregates data using a secret sharing scheme for privacy-preserving machine learning. However, their scheme has challenges with communication overhead. Other related works such as \cite{geyer2017, wei2020} have employed differential privacy to hide statistical information during multi-party training. However, UPs are not considered in their designs.

\section{Preliminaries}\label{sec:preliminaries}
\subsection{Homomorphic Encryption (HE)}\label{sec:he}
HE is a form of encryption that allows algebraic operations to be correctly performed on ciphertexts with the result remaining in an encrypted form \cite{gong2020}. Several HE schemes have already been proposed \cite{fontaine2007}, however, in this work we adopt the Paillier scheme \cite{paillier1999} which is an additive HE scheme. The scheme is proved to be secure and has been widely used in privacy-preserving multi-party machine learning works. We summarize its properties as follows:  

The Paillier scheme comprises three (3) algorithms: Key generation, Encryption, and Decryption algorithms. 
The Key generation algorithm generates the public and private keys. The public key is generated as ($n, g$), where $n=pq$, $p \text{ and q}$ are two large primes with $gcd(pq,(p-1)(q-1))=1$, and $g\in \mathbb{Z}^{*}_{n^2}$. Meanwhile, the private key is generated as ($\lambda, u$), where $\lambda=lmc(p-1, q-1)$ and $u=(L(g^\lambda \text{mod }n^2))^{-1}\text{mod }n$.

The Encryption algorithm is used to generate a ciphertext $C$. For a message $M \in \mathbb{Z}_n$, $C$ is produced as $C = g^M\times r^n \text{mod }n^2$, where $r\in \mathbb{Z}^{*}_{n^2}$.

The Decryption algorithm recovers $M$ from $C$. Given $C$, $M$ can be recovered as $M = L(C^\lambda \text{mod }n^2)\times u \text{ mod }n$. The details can be viewed in \cite{paillier1999}.

The Paillier scheme supports unlimited homomorphic addition operations and limited multiplication operations. Thus, an addition operation on two encrypted messages $M_1$ and $M_2$ results in an encrypted sum of the two messages, i.e.,
\begin{equation}
  E(M_1) + E(M_2) = E(M_1+M_2)
\end{equation}
where $E(.)$ represents an encryption operation. Also, an encrypted message $M$ raised to power $l$ produces an encrypted product of $M$ and $l$, i.e., 
\begin{equation}
  E(M)^l = E(M\times l)
\end{equation}
where, $l$ is an unencrypted constant.

\subsection{Machine Learning}
This section presents a brief review of the machine learning algorithms considered in this work: logistic regression and neural networks \cite{hastie2009}.

\subsubsection{Logistic Regression}
\begin{figure}[!t]
\begin{center}
\subfloat[]{\includegraphics[width = 1.5in]{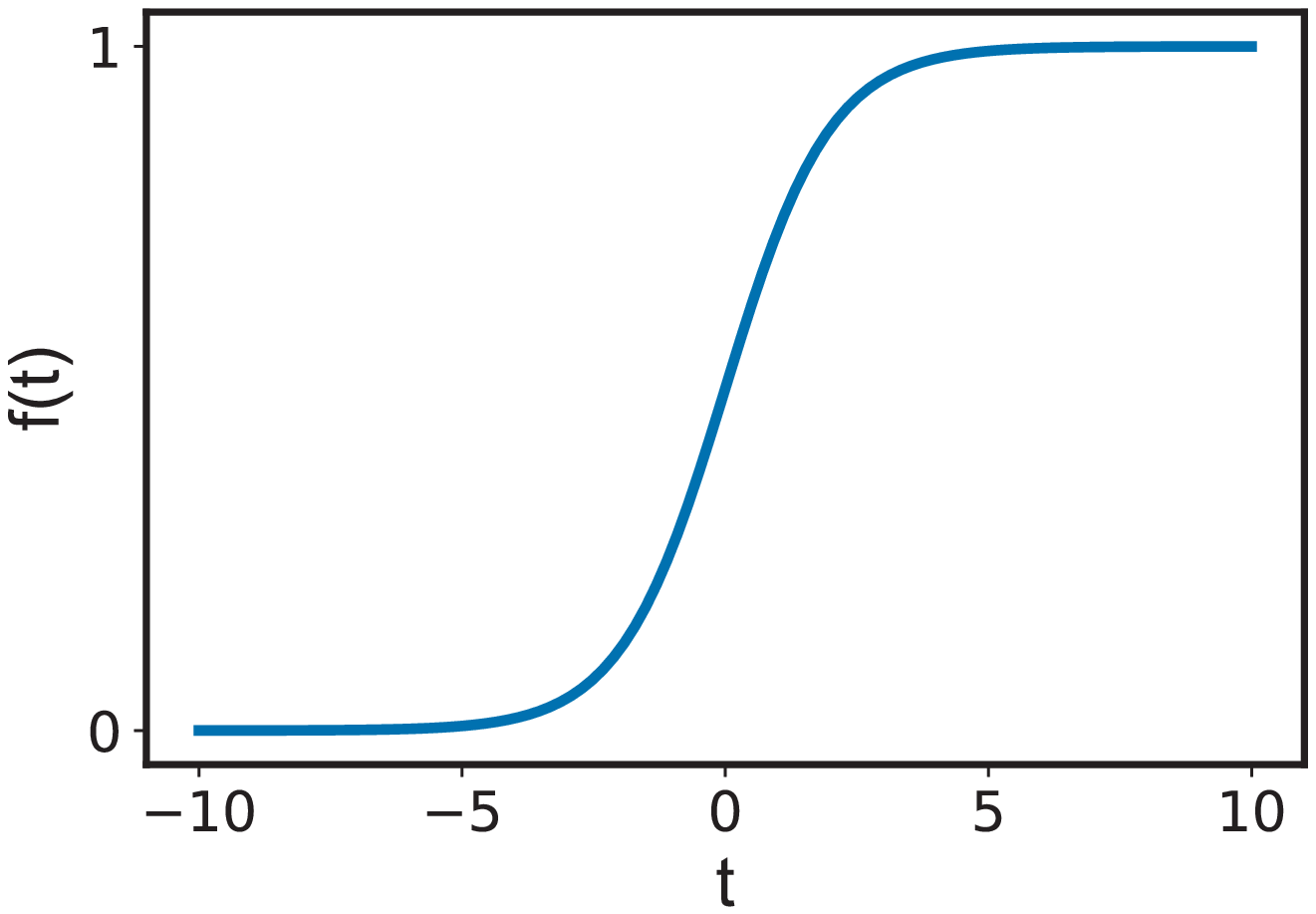}}
\subfloat[]{\includegraphics[width = 1.8in]{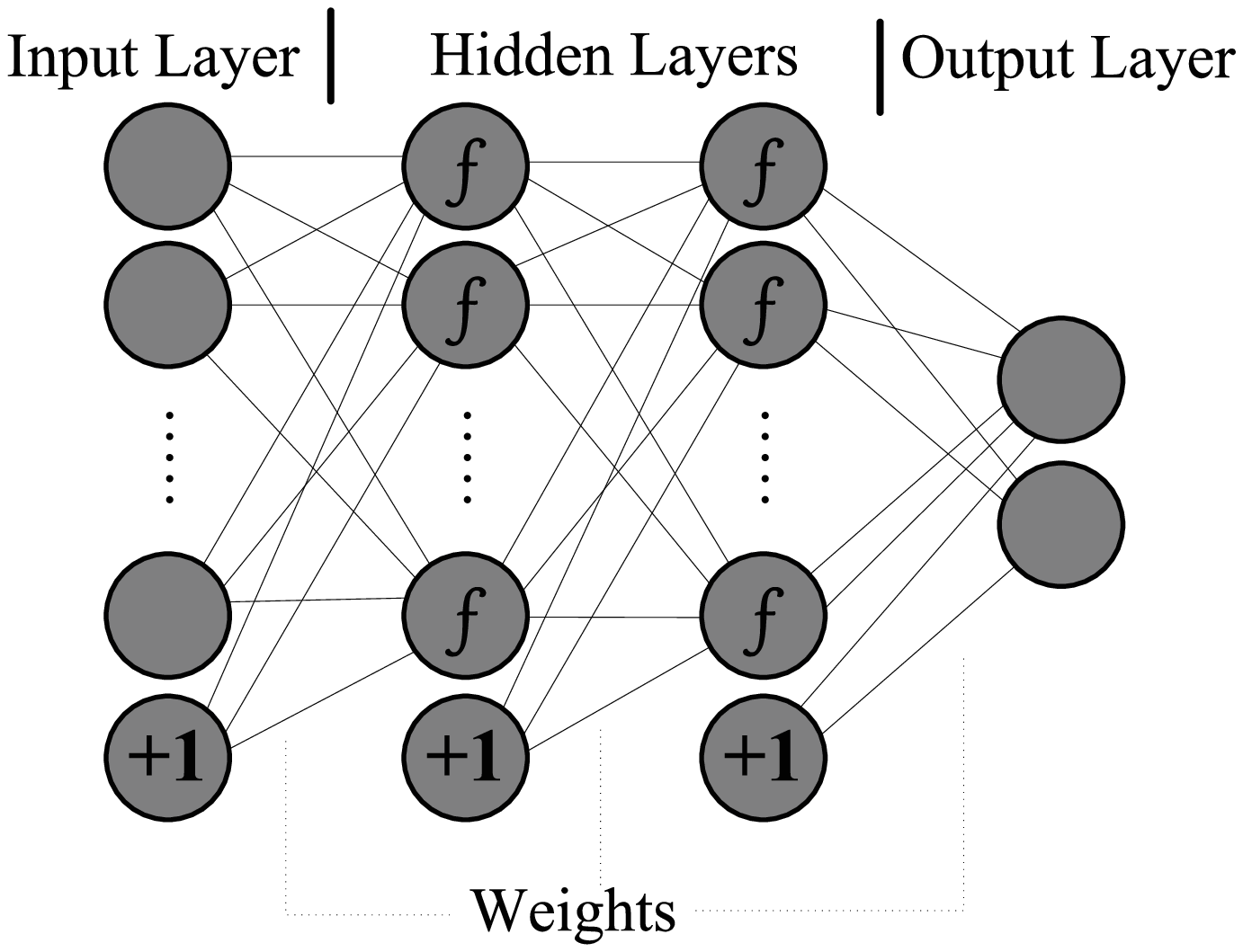}}
\caption{(a) A logistic activation function. (b) A neural network example.}
\label{fig:neural}
\end{center}
\end{figure}

Given a data item as $data = (x, y)$ with $x$ as the input and $y$ as the truth value, \textit{regression} is learning a function $g$ such that $g(x) \approx y$ \cite{mohassel2017}. In logistic regression, for a binary classification problem, the output value is bound between 0 and 1 through an activation function $f$. Therefore, the function $g(x)$ can be represented as $g(x) = f(x.W)$, where $W$ is the weight coefficient vector. The activation function $f$ used in logistic regression is defined as, $f(t) = \frac{1}{1+e^{-t}}$ which is shown in Figure \ref{fig:neural}(a). 
The cost function used in logistic regression is the cross-entropy function \cite{mohassel2017}. In this case, we simply write the cost function over the data item $data(x,y)$ as $C(W,x,y)$. 

\subsubsection{Neural Networks}
Neural networks generalize regressions to learn more complicated relationships in datasets. Figure \ref{fig:neural}(b) is an example neural network with four layers: an input layer, 2 hidden layers, and an output layer. Each node in the network is referred to as a \textit{neuron} and is associated with an activation function $f$. Examples of common activation functions used in neural networks are:
\[ \text{ReLU function } f(t) = max(0,t) \]
\[ 
\text{Leaky ReLU function } f(t) = 
\begin{cases}
  t & \text{if t $>$ 0}\\
  \beta t & \text{otherwise}\\
\end{cases}
\]

\[\text{tanh function } f(t) = \frac{e^t - e^{-t}}{e^t + e^{-t}}\]
\[\text{sigmoid function } f(t) = \frac{1}{1+e^{-t}}\]

The +1 node represents bias. The nodes in neural networks are connected through the weight vectors W. Neural networks have an additional function called the cost function. Common examples of cost functions are cross-entropy function, squared error cost function, etc \cite{phuong2019}. For a data item $data(x,y)$, we use the same representation $C(W, x, y)$ for the cost function in neural networks.

Thus, in any machine learning algorithm, the task is to determine the weight parameter that minimizes the cost function $C(W, x, y)$ for a given dataset.
%

\subsection{Stochastic Gradient Descent (SGD)}
SGD is an algorithm widely used in machine learning for approaching the global minimum of a function \cite{mohassel2017}. Given the weight parameters W, the SGD updates the parameters as:
\begin{equation}
 W:=W-\alpha.\frac{\delta C(W,x,y)}{\delta W} 
\end{equation}
where $\alpha \in \mathbb{R}$ is the learning rate.

In practice, for efficiency, instead of selecting a single data item at a time, multiple data items are selected inform of a matrix (X,Y). The matrix (X,Y) is referred to as a mini-batch. Thus, the parameter updates are based on mini-batches as:
\begin{equation}
 W:= W-\alpha.\frac{\delta C(W, X, Y)}{\delta W} 
\end{equation}

\subsection{Threat Model}
In our proposed system, we assume that the model initiator is honest-but-curious, i.e., it follows the steps the way they are but it might attempt to infer from the encrypted information. The server is also honest-but-curious and non-colluding, i.e., on top of being honest-but-curious, it does not collude with any participant to reveal information. And, the participants are malicious, i.e., a participant might attempt to infer another participant's private data or intentionally upload false parameters to the server.

\subsection{Similarity Computation}
Several similarity measurement techniques such as euclidean distance, jaccard similarity, cosine similarity, etc., have already been used in different machine learning algorithms. In this section, we review the cosine similarity which is of interest in this work. The cosine similarity between two F-dimensional vectors $U=\{u_1, \cdots, u_F\}$ and $V=\{v_1, \cdots, v_F\}$ is computed as:
\begin{equation}
  S_{cos}(U,V)=\sum_{i=1}^{F}\frac{u_i.v_i}{|U||V|}
\end{equation}
with $U \text{ and } V$ in plaintext form. In an encrypted form, the cosine similarity can be computed as:
\begin{equation}
  E(S_{cos}(U, V))= \prod_{i=1}^{F}E(\frac{u_i}{|U|})^{\frac{v_i}{|V|}}= \prod_{i=1}^{F}E(\frac{u_i}{|U|}.\frac{v_i}{|V|})
\end{equation}
using the properties of Paillier scheme discussed earlier. The cosine similarity computation outputs a value in the range [-1, 1], with 1 indicating total similarity between the two vectors and -1 indicating total dissimilarity between the vectors.

\section{Our Proposed Multi-Party Machine Learning System}\label{sec:our_work}
In this section, we present our proposed privacy-preserving multi-party machine learning system with reliability check. We discuss the system entities and their roles and provide security, efficiency and effectiveness analysis of the system elements. But first, we discuss the \textit{weight similarity} metric.

\subsection{Weight Similarity}\label{sec:weight}
For reliability check, we propose a new metric called the \textit{weight similarity} to measure the similarity between the participants' and the model initiator's datasets. To give a background on the \textit{weight similarity}, let us look at the following example. 

Consider two functions $f_1 = (pr-5)^2$ and $f_2 = (qr-5)^2$. The gradients of the two functions $f_1$ and $f_2$ can be computed as $\Delta_1 = 2p(pr-5)$ and $\Delta_2 = 2q(qr-5)$, respectively. 
Next, consider another function $f(r) = r-\gamma \Delta$. Substituting $\Delta$ with $\Delta_1$ and $\Delta_2$ gives us:
\begin{align}
\begin{split}
f(r)_1 = r-\gamma \Delta_1\\
f(r)_2 = r-\gamma \Delta_2
\end{split}
\end{align}
Therefore, if $p\approx q$ then $\Delta_1 \approx \Delta_2$ and $f(r)_1 \approx f(r)_2$. 

The above example depicts the relationship between the data items, gradients and weights in multi-party machine learning systems. The parameters $p$ and $q$ depict the data items of say two participants $A$ and $B$. The functions $f_1$ and $f_2$ are the cost functions of $A$ and $B$, respectively. Also, the gradients $\Delta_1$ and $\Delta_2$ depict the gradients generated by the participants $A$ and $B$, respectively. The function $f(r)$ depicts the weight update function of machine learning. Recall that, in a multi-party machine learning system, all the participants have the same model architecture. Thus, if the data items of the participants $A$ and $B$ are similar, there is a high likelihood that their generated gradients are similar. Thus, a parallel weight update by the participants $A$ and $B$ (i.e., a weight parameter updated by two different participants independently) using their respective gradients results in two similar weights. We exploit this property to identify RPs and UPs in our proposed system whose architecture is described in the next subsection. 


\subsection{System Architecture}
\begin{figure}[!t]
\begin{center}
\includegraphics[width = 4.0in]{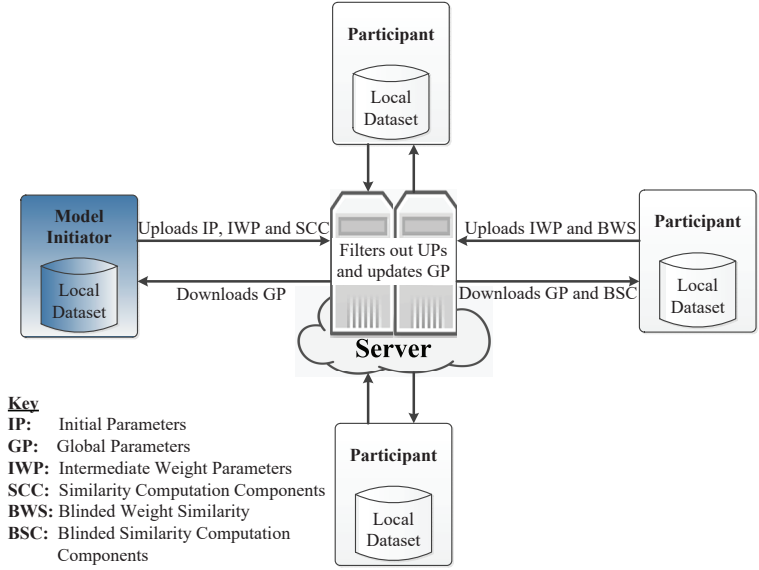}
\caption{A high-level architecture of our multi-party privacy-preserving machine learning with reliability check showing the involved entities.}
\label{fig:arch}
\end{center}
\end{figure}

The architecture of our proposed multi-party privacy-preserving machine learning with reliability check is depicted in Figure \ref{fig:arch}. The system consists of the following entities: a central \textit{server}, a \textit{model initiator}, and multiple \textit{participants} (which includes RPs and UPs). The model initiator collaborates with the server to initialize the system. Thereafter, the privacy-preserving learning process during which the model initiator and participants upload their intermediate weight parameters to the server begins. The server uses these intermediate weights to update the earlier set global parameters. During the update, the server uses weight similarity scores to filter out (exclude) the weight parameters from UPs. The weight similarity score is collaboratively computed by the entities in a privacy-preserving manner. The description of the entities is as follows:

\subsubsection{Server}
The central server stores encrypted global parameters and makes them available to the model initiator and the participants for download. It then receives encrypted weight parameters from both the model initiator and the participants. It also receives encrypted weight similarity computation components and blinded similarity scores from the model initiator and the participants, respectively. The server updates the global parameters with the received weight parameters. Depending on the similarity score and the set threshold value, the server might include (or exclude) a participant's weight parameters during the global parameter updates, i.e., the server filters out weight parameters from UPs when updating the global parameters. We assume the server to be \textit{honest-but-curious}, i.e., it follows the algorithm the way it is, but it is curious about the data.

\subsubsection{Model Initiator and Participants}
The model initiator is an RP who sets the initial global parameters. 
The model initiator and the participants aim to learn a common model and thus, they share an identical model architecture and learning objective. They also share a homomorphic private key that is kept secret from the server. Therefore, E(.) depicts a homomorphic encryption operation, which protects the privacy of the exchanged parameters. The model initiator and the participants each keep their local datasets but only exchange encrypted intermediate parameters with the server. The exchanges happen at every communication round, which is decided by the server. For example, the server might schedule the model initiator and the participants to upload their intermediate parameters after every 10 local epochs. Thus, the training is done synchronously.

\begin{figure}[!t]
\centering
\includegraphics[width=4.0in]{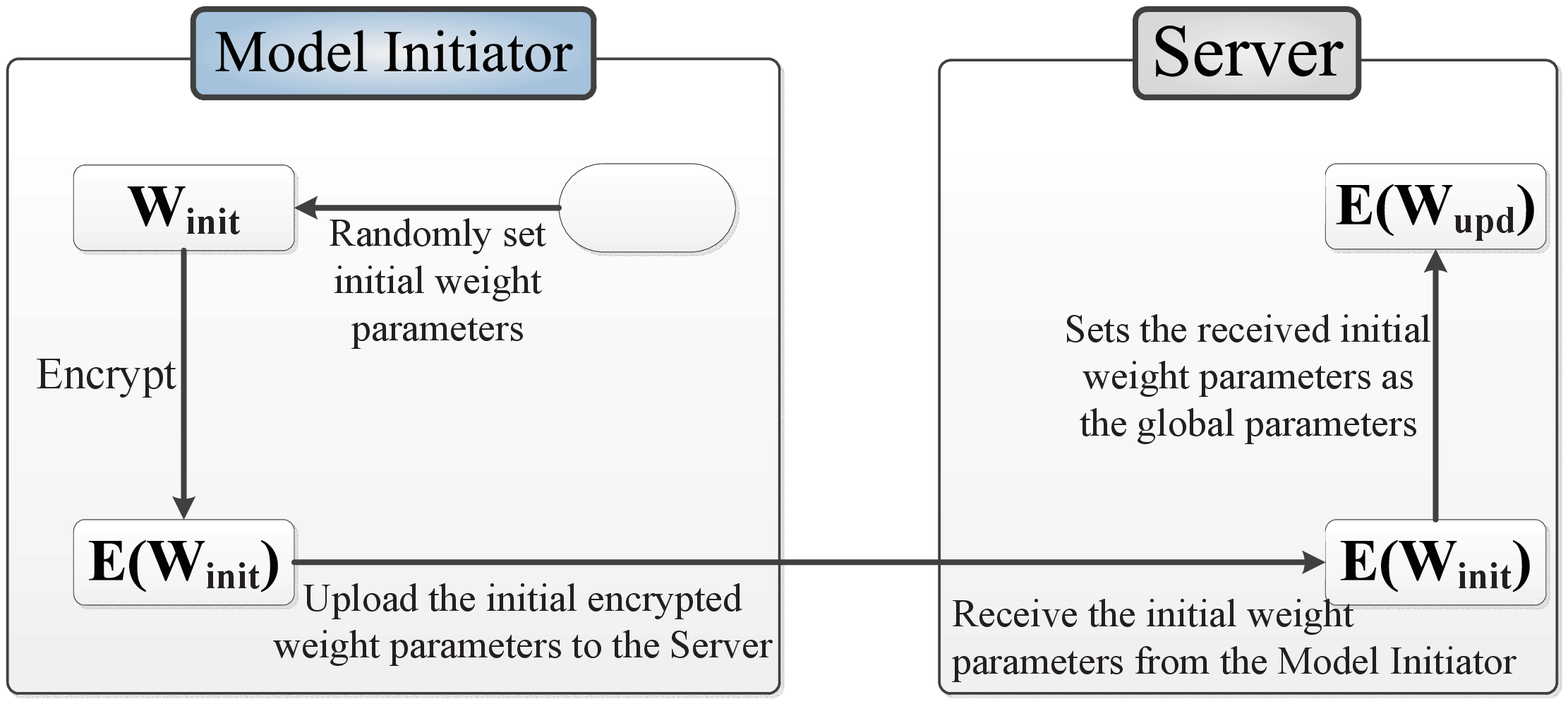}
\caption{Interaction between the model initiator and the server during the initialization phase.}
\label{fig:init}
\end{figure}

The model initiator runs two phases (\textit{initialization} phase and \textit{learning} phase), while the participants run only one phase (\textit{learning} phase)\footnote{However, the model initiator's and the participants' learning phases are not entirely identical.}. During the initialization phase, the model initiator sets the initial parameters $W_{init}$ which it encrypts as E($W_{init}$) and sends them to the server. Upon reception, the server sets E($W_{init}$) as the global parameters E($W_{upd}$) and makes them available to the participants for download. The process is illustrated in Figure \ref{fig:init}.

The model initiator and the participants then enter the learning phase and perform local training with their private datasets using the SGD and upload their encrypted intermediate weight parameters to the server at every communication round. The model initiator and the participants compute an additional component used by the server to establish a weight similarity score for each participant's uploaded weights. The server uses the similarity score to determine if a participant's parameters should be included in the global parameter updates. Next, the server updates the global parameters and makes them available to the participants, and the model initiator for the local training to continue. 
We present the detailed procedures in subsequent sections.

\subsection{The Model Initiator Side Procedure}

\begin{figure}[!t]
\centering
\includegraphics[width=6.0in]{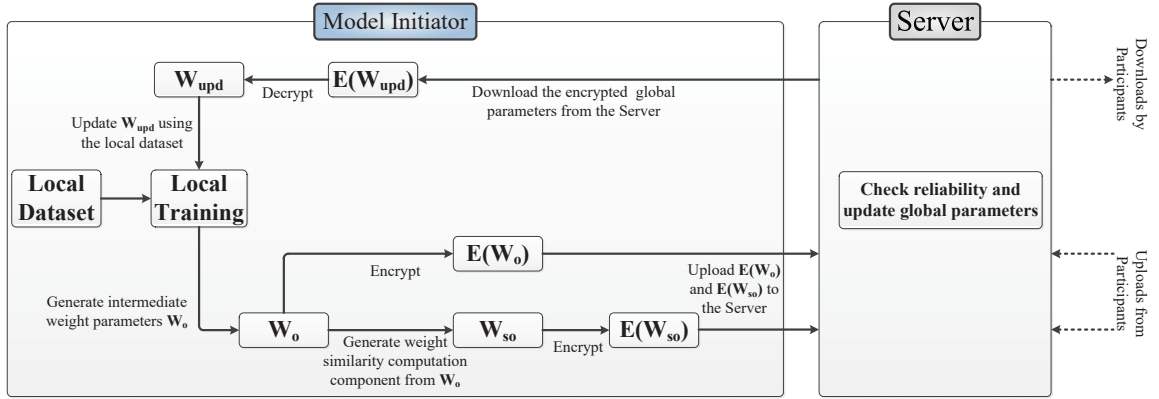}
\caption{Learning process of the Model Initiator showing its interactions with the Server.}
\label{fig:mi_learning}
\end{figure}

We show the pseudocode executed on the model initiator side in Algorithm \ref{alg:initiator}. The model initiator has its own local dataset for conducting training using the standard SGD, however, the dataset is insufficient. As stated in the previous subsection, first, the model initiator initializes the weight parameters as $W_{init}$. It then encrypts the parameters using the HE algorithm discussed in section \ref{sec:he} as E($W_{init}$) and sends them to the server from where they are globalized as E($W_{upd}$) and made available for download by the participants. Note that the initialization phase is executed once.   

Starting from $W_{init}$, the model initiator then trains its local model by running the standard SGD on its insufficient local dataset to generate the intermediate weight parameter $W_o$, which it encrypts as E($W_o$). It also generates a weight similarity computation component $W_{so}$\footnote{$W_{so}= W_{o}^{i}/|W_o|_{\forall i}, \text{where}, W_{o}^{i} \in W_o$.} which it encrypts as E($W_{so}$) and forms the base part of Equation 6. The model initiator sends E($W_o$) and E($W_{so}$) to the server at each communication round. 

Next, the model initiator downloads the updated global parameters E($W_{upd}$) from the server during the subsequent communication rounds, decrypts them using its private key and continues with the training process using its local dataset until the model improvement is minimal and all the participants have stopped. A diagrammatic illustration of the learning process (especially after the first communication round) is presented in Figure \ref{fig:mi_learning}.

\subsubsection{Security Analysis I}
\textit{Theorem 1 (Security against an honest-but-curious model initiator):} An honest-but-curious model initiator learns no information about the participants' private data from the received global parameters in Algorithm \ref{alg:initiator} (line 5). 

\textit{Proof:} The updated global parameters are computed from the intermediate weight parameters of the participants and thus, an honest-but-curious model initiator learns no information about the private data of the participants. 

\textit{Remark 1 (Regarding the privacy-preservation via weights):} In \cite{phuong2019}, Phong and Phuong proved that participants' private data cannot be retrieved from intermediate weight parameters. Therefore, since the model initiator only receives the global parameters computed from the intermediate weights of the participants, it cannot retrieve the private data of any participant.  

\begin{figure}[!t]
\removelatexerror
\begin{algorithm}[H]
\DontPrintSemicolon
\SetAlgoLined
 
1: Initialize the parameters as $W_{init}$\\
2: Encrypt the initial parameters as E($W_{init}$) and send them to the server\\
3: Download global parameters E($W_{upd}$) from the server and decrypt it as $W_{upd}$ (except at the beginning when $W_{init}$ is used), perform local SGD based on the local dataset to generate the parameters $W_o$ and $W_{so}$\\
4: Encrypt $W_o$ as E($W_o$) and $W_{so}$ as E($W_{so}$), and send them to the server\\
5: Receive updated parameters E($W_{upd}$) from the server and decrypt them as $W_{upd}$\\
6: Repeat steps 3-5 until the validation loss is acceptably small and the other participants have dropped out of the training\\
7: Stop the training process
    
    
 
%
%
\caption{Model Initiator Side}\label{alg:initiator}
\end{algorithm}
\end{figure}

\subsection{The Server Side Procedure}
The pseudocode of our scheme on the server side is shown in Algorithm \ref{alg:server}. As mentioned in the preceding subsections, the server first receives the initial weight parameters E($W_{init}$) from the model initiator. It then sets E($W_{init}$) as the global parameter E($W_{upd}$) which it makes available to the participants for download. At each communication round, the server receives the intermediate parameters E($W_o$)  and E($W_{so}$) from the model initiator. E($W_o$) is used during the global parameter update while E($W_{so}$) is used for secure weight similarity score computation. 

The server then initiates a secure computation of the weight similarity score between the model initiator's and the participants' intermediate weight parameters. To achieve this, the server first blinds E($W_{so}$) by computing E($W_{so}$)$^l$ = E($W_{so} l$), where $l \in \mathbb{R}_{\neq (0,1)}$. It then sends E($W_{so} l$) to the participants. The server waits for the participants computation and then receives an encrypted intermediate weight parameter E($W_p$) and a blinded weight similarity score $S_{cos} l$ from each participant.


The server then computes the final weight similarity score as:
\begin{equation}
  S_{cos}(W_o, W_p) = \frac{S_{cos} l}{l}
\end{equation}
After computing the weight similarity scores for all the participants, the server updates the global parameters E($W_{upd}$) by averaging using Equation 9. 
\begin{equation}
  \text{E($W_{upd}$)} = \big(\text{E($W_o$)}+\sum_{p=1}^{P}\text{E($W_p$)}\big)^{(\frac{1}{P+1})}
\end{equation}
where $P$ is the number of participants whose weight similarity scores are above the threshold value $T$, i.e., E($W_p$) is only included in the parameter update if and only if the following condition holds for its corresponding weight similarity score, 
\begin{equation}
  T< S_{cos}(W_o, W_p)\leq 1
\end{equation}
where, $T$ is a threshold value, which can be fixed or increased dynamically at each communication round.
The server finally makes the updated global parameters available for download by the participants and the model initiator to continue with their training processes. 

\begin{figure}[!t]
\removelatexerror
\begin{algorithm}[H]
\DontPrintSemicolon
\SetAlgoLined
1: Receive the initial parameters E($W_{init}$) from the model initiator, sets E($W_{init}$) as E($W_{upd}$) and make them available to the participants\\
2: Receive E($W_o$) and E($W_{so}$) from the model initiator\\
3: Compute (E($W_{so}$)$^{l}$=E($W_{so} l$), where $l \in \mathbb{R}_{\neq (0,1)}$\\
4: Send E($W_{so} l$) to the participants\\
5: Receive $S_{cos} l$ and E($W_p$) from each participant\\
6: Compute similarity score for each participant as $S_{cos}(W_o,W_p)=\frac{S_{cos} l}{l}$\\
7: \textbf{Forall }{$S_{cos}(W_o,W_p)> T$} \textbf{ then}\\
  \hspace{0.5cm}Compute E($W_{add}$)=E($W_o$)$+$E($W_1$)$+$$\cdots + $E($W_P$)=E($W_o+W_1 + \cdots + W_P$)\\
  \hspace{0.4cm}\textbf{end Forall}\\
\BlankLine
8: Update the global parameters as E($W_{upd}$)=E($W_{add}$)$^{(P+1)^{-1}}$, where $P$ is the number of participants whose weight similarity score with the model initiator's is above the threshold $T$\\
9: Send E($W_{upd}$) to all the participants including the model initiator\\
10: Repeat steps 2-9 until the model initiator stops the training process
\caption{Server Side}\label{alg:server}
\end{algorithm}
\end{figure}

\subsubsection{Security Analysis II}
\textit{Theorem 2 (Security against an honest-but-curious server):} An honest-but-curious server learns no information about the trained model and the private datasets used in the training.

\textit{Proof:} An honest-but-curious server only computes on the encrypted parameters from the model initiator and the participants. Therefore, it obtains no information about the model, and the local datasets of the model initiator and the participants since the encryption scheme used is secure. 

\textit{Remark 2 (Regarding the Computation of Weight Similarity Score):} The computation of weight similarity scores aims at identifying RPs. The computation involves an exchange of encrypted intermediate parameters between the model initiator and the participants through the server. Since the server does not have access to the private key, the exchanged intermediate parameters are kept secure from the server. It is only the final weight similarity score that gets revealed to the server.


\subsection{The Participant Side Procedure}
Shown in Algorithm \ref{alg:participant} is the pseudocode of our scheme the participants execute. Like with the model initiator, each participant runs the standard SGD on its own local dataset. Each participant first downloads the global parameter E($W_{upd}$) from the server and decrypts it with the shared homomorphic private key. Each participant then runs the standard SGD on its local training dataset to generate the intermediate weight parameter $W_p$ at each communication round. 

To compute the weight similarity score, each participant receives E($W_{so} l$) from the server. Next, each participant generates an encrypted and blinded weight similarity score $E(S_{cos} l)$ by computing $\prod_{i=1}^{F} E(W_{so}l^{i}.W_{sp}^{i})$ according to Equation 6\footnote{Where, $W_{sp}^{i}$ which forms the power part of Equation 6 is an element of $W_{sp}$, $W_{sp}=W_{p}^{i}/|W_p|_{\forall i}$ is the similarity computation component of a participant and $W_{p}^{i}\in W_{p}$.}. Each participant then encrypts $W_p$ as E($W_p$) and decrypts E($S_{cos} l$) as $S_{cos} l$. Next, each participant sends E($W_p$) and $S_{cos} l$ to the server. 

\begin{figure}[!t]
\centering
\includegraphics[width=6.5in]{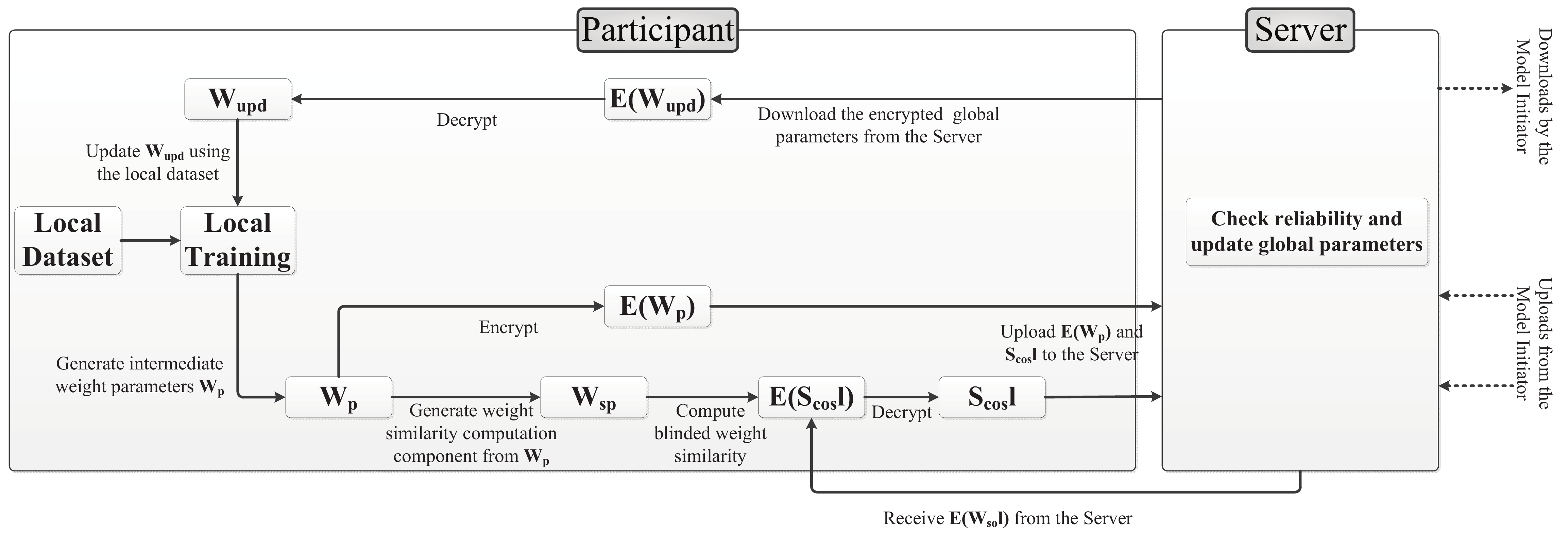}
\caption{Learning process of a participant showing its interactions with the Server.}
\label{fig:p_learning}
\end{figure}

Each participant then waits for the server to update the global parameters and then downloads the updated global parameters E($W_{upd}$) to continue with the training process using its local dataset. This is repeated until the accuracy improvement is minimal. However, a participant can decide to quit at any time. An illustration of the process is shown in Figure \ref{fig:p_learning}.

\begin{figure}[!t]
\removelatexerror
\begin{algorithm}[H]
\DontPrintSemicolon
\SetAlgoLined
1: Download the global parameters E($W_{upd}$) from the server and decrypt it as $W_{upd}$\\
2: Perform local SGD based on the local dataset to generate $W_p$\\
3: Receive E($W_{so} l$) from the server and compute E($S_{cos} l$)=$\prod_{i=1}^{F} E(W_{so}l^i. W_{sp}^{i})$\\
4: Encrypt $W_p$ as E($W_p$)\\
5: Decrypt E($S_{cos} l$) as $S_{cos} l$\\
6: Send E($W_p$) and $S_{cos} l$ to the server\\
7: Repeat steps 1-6 until the validation is acceptably small\\
8: Drop out of the training process
\caption{Participant Side}\label{alg:participant}
\end{algorithm}
\end{figure}

\subsubsection{Security Analysis III}
\textit{Theorem 3 (Security against a malicious participant):} A malicious participant learns no information about the model initiator's parameters during the weight similarity score computation procedure in Algorithm \ref{alg:participant} (line 3). Also, a malicious participant learns no information about the private data of the model initiator and the other participants from the updated global parameters received from the server. 

\textit{Proof:} The similarity computation component forwarded to the participant is blinded by the server with a non-zero random value in our system. Therefore, a malicious participant obtains no information regarding the true parameter values of the model initiator. The proof for the data privacy of the model initiator and the other participants is similar to the one of the model initiator procedure. 

\textit{Remark 3 (Regarding Computation of the Weight Similarity Score Component by the Participant):} The computation and decryption of the blinded weight similarity scores by the participants aim at enabling the server to securely update the global parameters using intermediate weight parameters from only RPs without having access to the private key. However, this comes with additional computation and communication overhead. We leave the question of revealing the weight similarity score to the server with minimal computation and communication overhead open for future considerations.

\subsection{Effectiveness and Efficiency Analysis}
\subsubsection{Effectiveness}
Here, we analyze the effectiveness of our proposed metric, which minimizes the disruptions caused by participants with noise data during privacy-preserving multi-party machine learning. Generally, the training of a machine learning model is a process for fine-tuning weights from random weights. The fine-tuning is guided by gradients that drive the learning process towards the local optimal solutions. In similar datasets, these gradients are similar and the update directions of the weights are almost the same. However, in the presence of noise, some update directions might be reversed, which contributes to weight similarities and dissimilarities between weight parameters of multiple participants in collaborative machine learning. Thus, setting a suitable weight similarity threshold from the beginning of the training minimizes the training disruption that would arise from the noise data of UPs. This threshold value can be fixed or dynamically raised as the learning tends towards the optimal solution. Our proposed approach does not guarantee any accuracy improvement on centralized training in which the dataset from all the participants is gathered centrally for training a model, but it guarantees improved convergence and reduced inaccuracies.

\subsubsection{Efficiency}
The efficiency of our proposed system can be analyzed from four perspectives. From the perspective of the model initiator, the privacy-preserving operation, and the similarity computation component and its encryption can be designed to run in parallel especially after the generation of the intermediate weight parameters to speed up the process. Thus, the similarity computation operation has a reduced impact on the training efficiency of the model initiator. 

From the perspective of the server, we employ the additively homomorphic Paillier algorithm that supports only addition operations with limited multiplication operations, and thus, it is more efficient as compared to those that perform arbitrary algebraic operations.

From the perspective of the participants, similar to the model initiator, the privacy-preserving operation, and the weight similarity score generation and its decryption can be performed in parallel after the generation of the intermediate weight parameters. This minimizes the impact of the weight similarity computation on the training efficiency of the participants. 

From the general view, all the participants and the model initiator can perform their local training using their local datasets simultaneously, which is equivalent to data parallelism. The main computational overhead only arises from the privacy-preserving operation using the Paillier algorithm. The computation demand for weight similarity score generation can be minimized through parallel computations.

\section{Experiments}\label{sec:experiments}
To demonstrate the applicability of our proposed system, in this section, we present the performance evaluation of our scheme with experiments on real-world datasets (MNIST \cite{mnist} and CIFAR-10 \cite{cifar}). We employ a desktop computer with an Intel(R) Core(TM) i5-6500 3.20GHz CPU, a GeForce GT 710 GPU, and 16GB RAM, running on Ubuntu 20.04 operating system for all the experiments. The Paillier algorithm library in \cite{PythonPaillier} with key length of 1024 bits is used in the experiment. 

We mainly compare our scheme with the Phung and Phuong scheme \cite{phuong2019} (referred to as the PP system for simplicity in this section), and two baselines (\textit{centralized} and \textit{stand-alone}). In the \textit{centralized} baseline, the model is trained on a centralized good quality dataset and is ought to achieve the best performance. Since there is no collaboration between different participants, no privacy-preserving mechanism is included during the training. Meanwhile, in the \textit{stand-alone} baseline, the model is trained using only the free-noise local dataset. And since there is no collaboration involved, no privacy-preserving mechanism is considered during the model training as well. A reliability check is not considered in the PP system \cite{phuong2019}. Thus, during the experimentation, weight parameters from all the participants are considered indiscriminately. 

There are two categories of participants simulated in our proposed system, i.e., \textit{reliable participants} (RP) and \textit{unreliable participants} (UP). The local dataset of a RP is similar to the local dataset of the model initiator which is noise-free. Meanwhile, the local dataset of a UP is noisy, i.e., a fraction of UP's local dataset is filled with noise. In this case, all the participants and the model initiator execute the same model architecture.

\subsection{Experiments with the MNIST Dataset}
In this section, we perform experiments with the MNIST dataset using a logistic regression model and a multi-layer perceptron (MLP) model. 

\subsubsection{Datasets}
The MNIST dataset comprises 28 x 28 gray-scale handwritten digit images with a training set of 60,000 images and a test set of 10,000 images. In our experiment, we simulate four (4) RPs, two (2) UPs, and a model initiator, and the training and test set images are divided amongst the participants and the model initiator. Each RP is allocated 10,000 samples of the training set and 1,660 samples of the test set. Similarly, the model initiator is allocated 10,000 samples of the training set and 1,660 samples of the test set. For the UPs, each participant's training set comprises 5,000 samples of the MNIST's training set and 5,000 samples of noise data, and the test set comprises 850 samples of MNIST's test set and 810 samples of noise data. In this case, we employ the notMINST dataset \cite{notmnist} as the noise data. The notMNIST dataset consists of gray-scale images of "A" to "J" letters formatted as 28 x 28 images, with a training set of 500,000 images and a test set of 19,000 images. The noise data of 5,000 for the training set and 810 for the test set for each UP is randomly sampled from the notMNIST's training and test sets, respectively. All the images are normalized and centered during the experiment.

\subsubsection{Using a Logistic Regression Model} 
We implemented a logistic regression model in Python using Theano 1.0.4. We set the random seeds as \\\textit{numpy.random.seed(139)} and \textit{random.seed(1234)}. The SGD is used with a fixed learning rate of 0.13 and a batch size of 128. All the participants and the model initiator run the same code. In a communication round, each participant and the model initiator runs 5 local epochs before encrypting and uploading intermediate parameters to the server. We set a fixed weight similarity score threshold of 0.05 during the training. 

\begin{figure}[!t]
\begin{center}
\subfloat[]{\includegraphics[width = 2.2in]{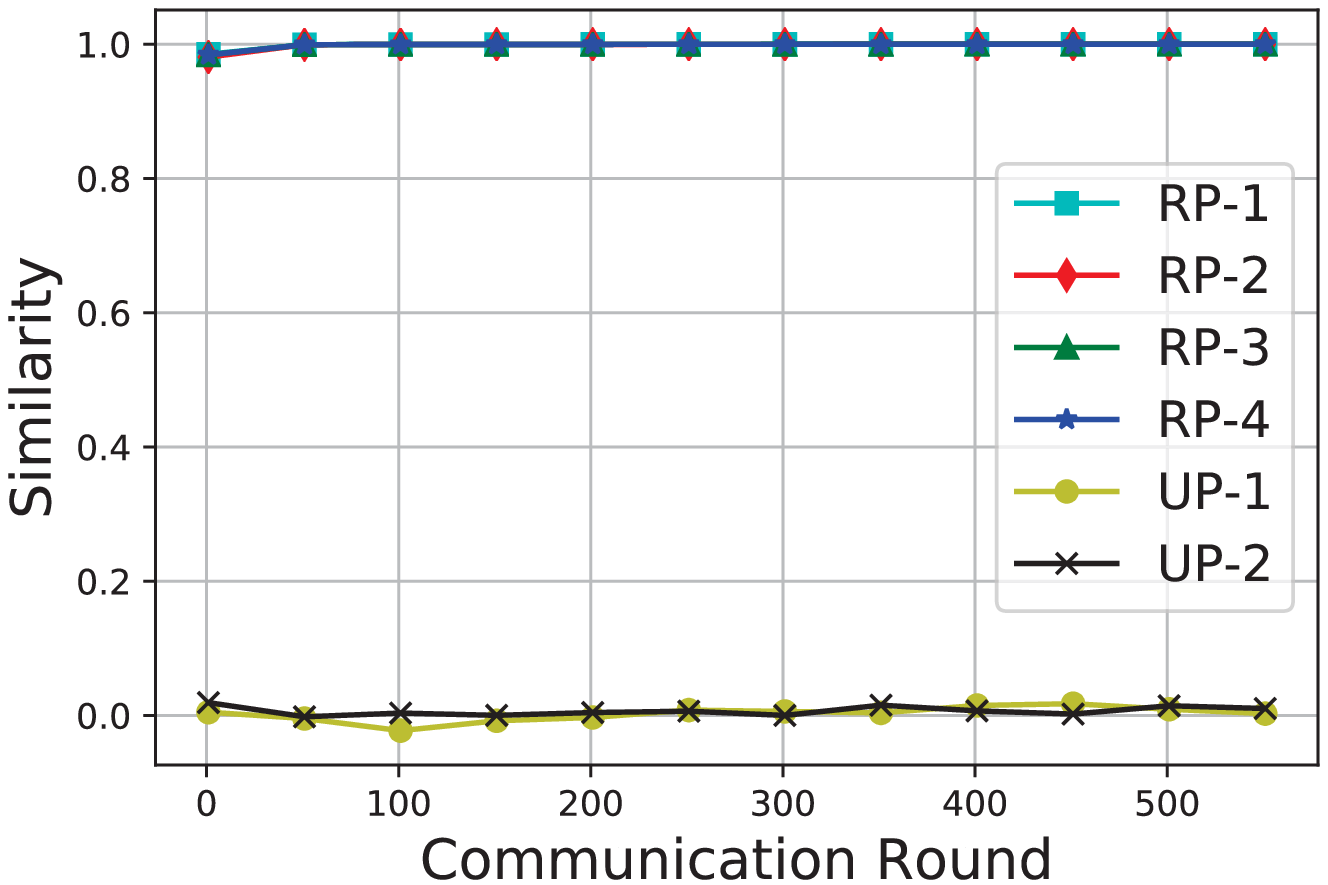}}
\subfloat[]{\includegraphics[width = 2.2in]{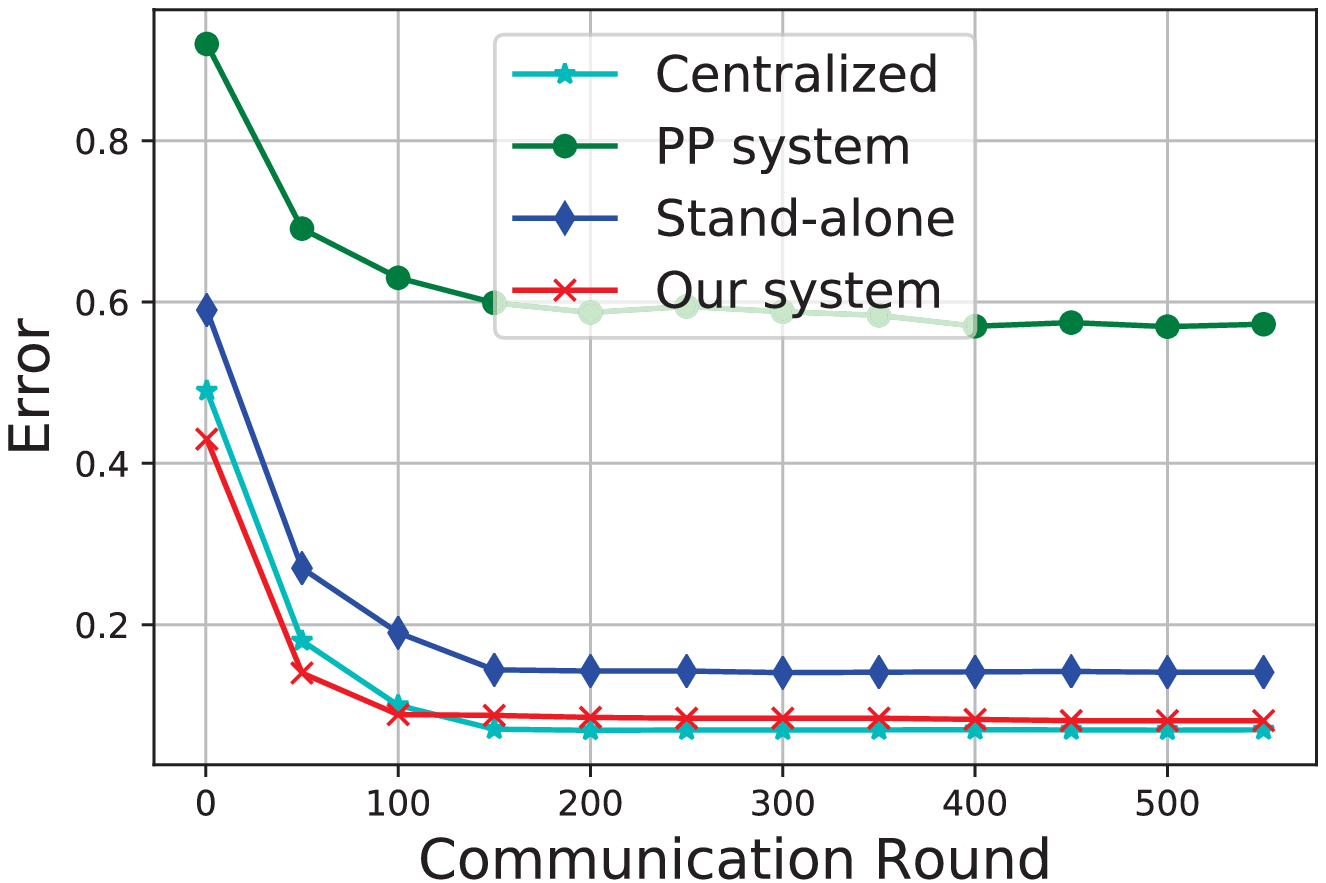}}
\subfloat[]{\includegraphics[width = 2.2in]{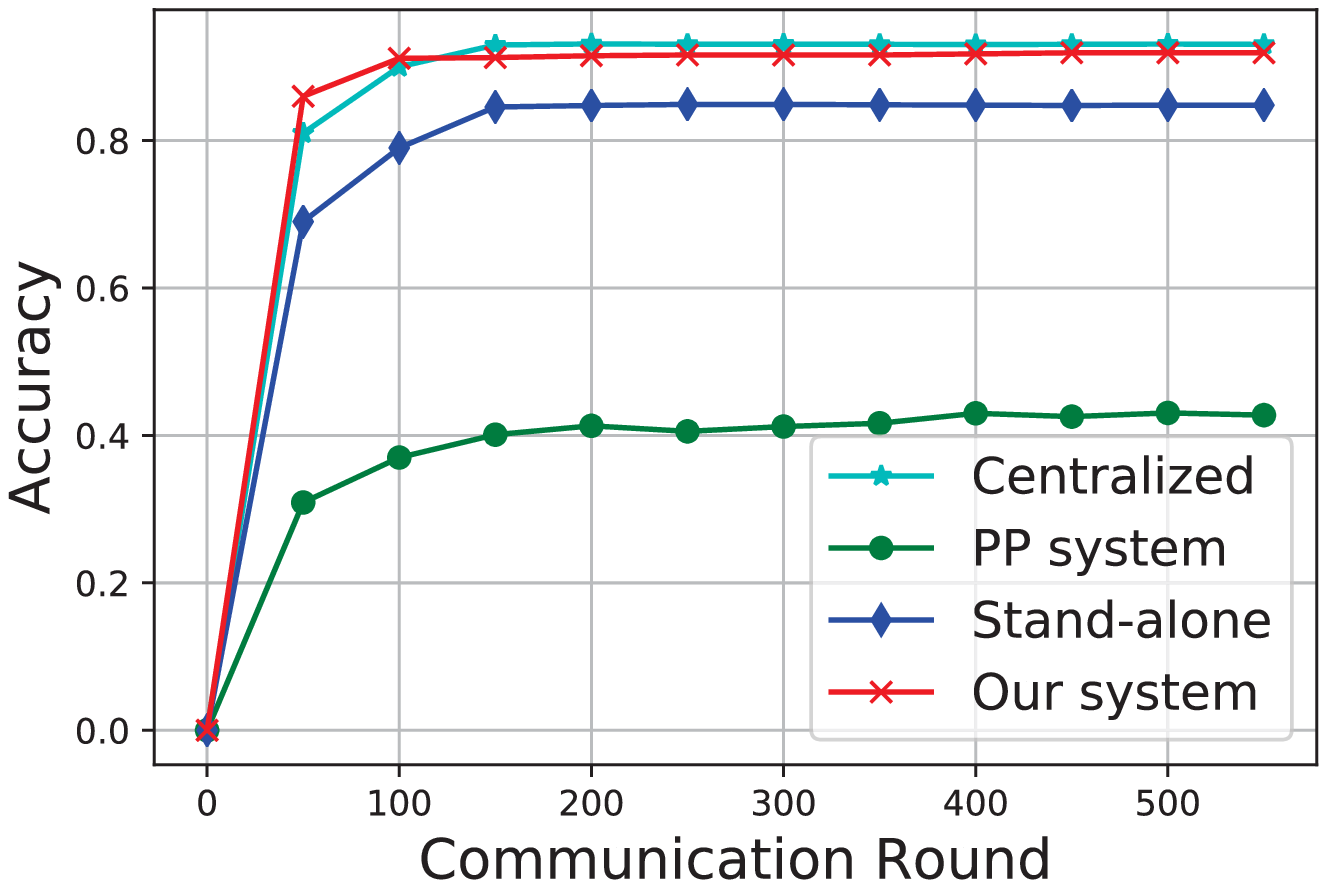}}\\
\caption{Results of logistic regression with MNIST dataset. (a) Weight similarities of the participants' weights to the model initiator's weight. (b) Test errors against communication round. (c) Accuracies against communication round.}
\label{fig:lr_mnist}
\end{center}
\end{figure}

\textit{Results:} The results of using the logistic regression model are shown in Figure \ref{fig:lr_mnist}. Figure \ref{fig:lr_mnist}(a) demonstrates the similarity between the participants' and the model initiator's weight parameters. The model initiator's and the RPs' weight parameters are highly similar, while the model initiator's and the UPs' weight parameters are less similar. Figure \ref{fig:lr_mnist}(b) and Figure \ref{fig:lr_mnist}(c) depict the training convergence against the number of communication rounds. Our proposed scheme achieves faster convergence as compared to the baseline schemes and the PP system which indiscriminately combines weight parameters from all the participants including the UPs during the global parameter updates. On the contrary, our system only considers weight parameters from RPs during the global parameter updates, i.e., it only uses the weight parameters from participants whose weight similarity score with the model initiator is above the threshold. 

\begin{figure}[!t]
\begin{center}
\subfloat[No. of local epochs = 5]{\includegraphics[width = 2.2in]{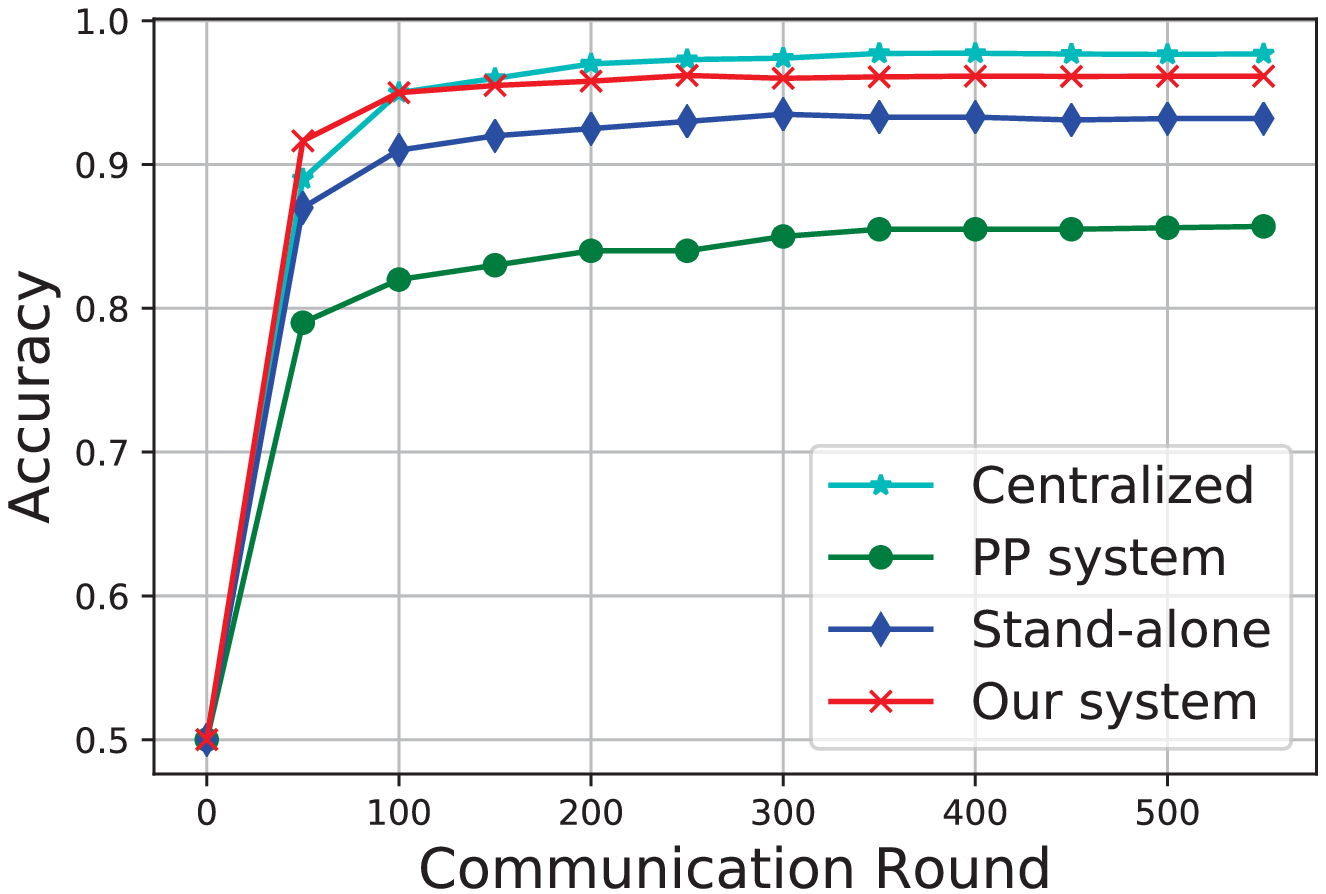}}
\subfloat[No. of local epochs = 10]{\includegraphics[width = 2.2in]{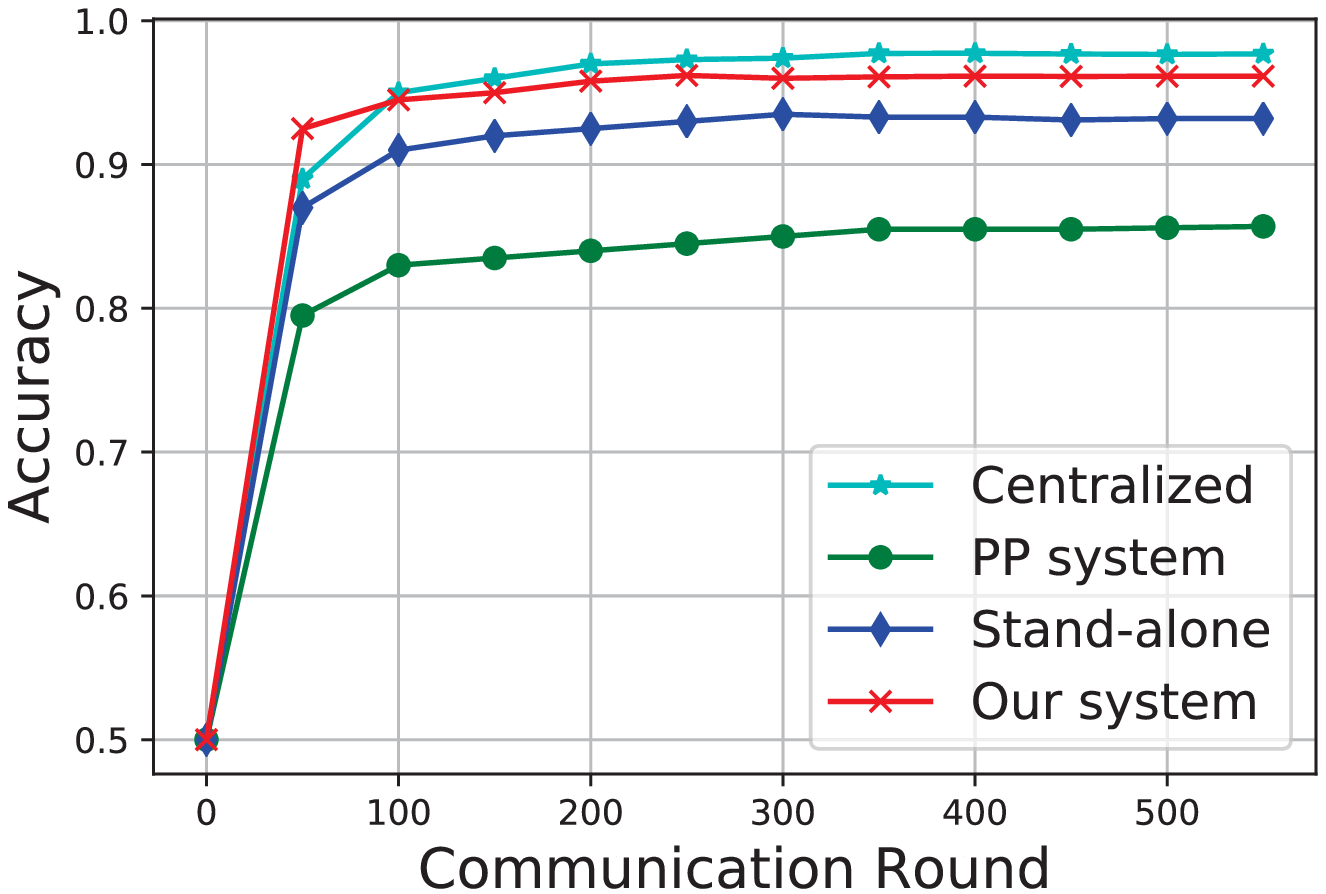}}
\subfloat[No. of local epochs = 50]{\includegraphics[width = 2.2in]{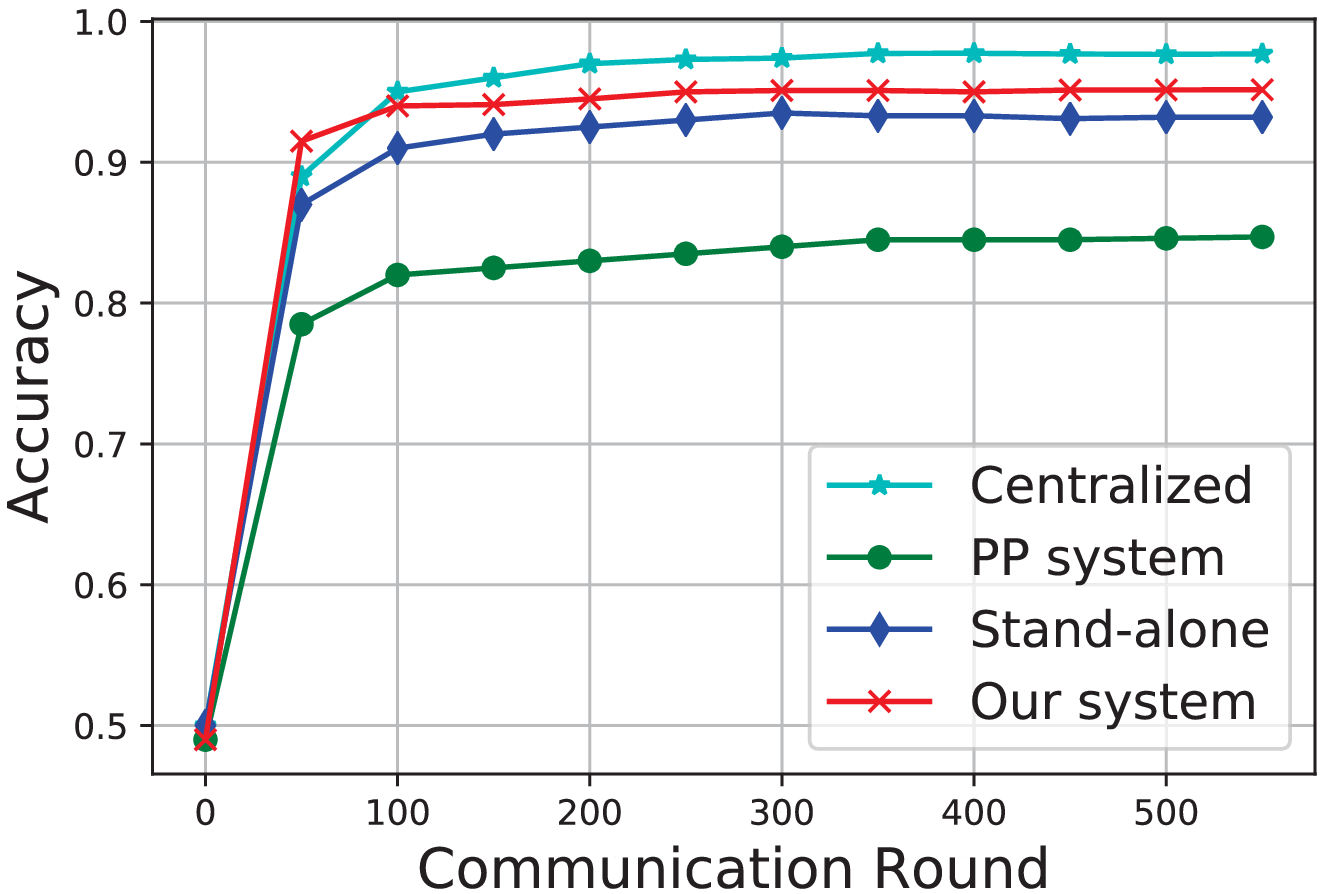}}\\
\caption{Results of MLP with MNIST dataset showing convergence against the number of communication rounds at different number of local epochs.}
\label{fig:nna_mnist}
\end{center}
\end{figure}

\subsubsection{Using an MLP} 
We also implemented an MLP model in Python and TensorFlow 2.1.0 with two hidden layers each with 64 neurons. We used ReLU as the activation function for the hidden layers and Sigmoid for the output layer. We set the random seeds as \textit{numpy.random.seed(12)}, \textit{random.seed(1234)} and \textit{tensorflow.set\_random\_seed(12345)}. We used the SGD with a fixed learning rate of $10^{-3}$ and a batch size of 64. All the participants and the model initiator run the same code. In each communication round, we vary the number of epochs in the local training. We also dynamically varied the similarity score in our experiment in the range of 0.1-0.7 in intervals of 0.1 after every 100 communication rounds. This is because, as the model tends towards convergence, the parameters from RPs become more similar.

\textit{Results:} The results of using an MLP are shown in Figure \ref{fig:nna_mnist}. Our proposed system converges faster than the baseline schemes and the PP system. However, when the number of local epochs is increased to 50, there is a slight drop in the accuracy of our system. This could be due to the improper mixing of parameters. The PP system which does not filter out UPs converges slowly and it is the least accurate. The centralized baseline achieves the best accuracy as expected. However, the stand-alone baseline is slightly less accurate because of the inadequate amount of training data items. In terms of accuracy, our scheme is only slightly bettered by the centralized baseline scheme.



\subsection{Experiments with the CIFAR-10 Dataset}
We also perform experiments with the CIFAR-10 dataset using a logistic regression model and an MLP model.

\subsubsection{Datasets}
The CIFAR-10 dataset consists of 60,000 RGB images of 10 different classes formatted as 32 x 32 x 3. 50,000 of these images form the training set while 10,000 of the images form the test set. In this experiment, we simulate three (3) RPs, a single UP, and a model initiator. Each RP holds 11,000 samples of the CIFAR-10 training set as its local training set and 2,200 samples of the CIFAR-10 test set as its local test set. Similarly, the model initiator holds 11,000 samples of the CIFAR-10 training set as its local training set and 2,200 samples of the CIFAR-10 test set as its local test set. The UP holds the remaining 6,000 samples of the CIFAR-10 training set and 5,000 samples of a noise dataset as its local training set, and 1,200 samples of the CIFAR-10 test set and 1,000 samples of a noise dataset as its local test set. As in the first case, here, we employ the notMNIST dataset as the noise data. However, notMNIST dataset is formatted as 28 x 28, thus, to correctly use it with the CIFAR-10 dataset, we padded it with zeros to obtain the same dimensionality as that of CIFAR-10. Therefore, the noise data of 5,000 and 1,000 for training and test sets are extracted from the padded notMNIST dataset. All the images are normalized and centered.

\subsubsection{Using a Logistic Regression Model}
Using Python and Theano 1.0.4, we implemented a logistic regression model to demonstrate the applicability of our system on the CIFAR-10 dataset. In the experiment, random seeds are set as \textit{numpy.random.seed(15)} and \textit{random.seed(123)}. We used the SGD in the learning process with a fixed learning rate of 0.01 and a batch size of 64 data items. All the participants and the model initiator execute the same model on their local datasets and upload their encrypted intermediate results to the server after every 5 local epochs. We set a fixed similarity score threshold of 0.03 for global parameter updates in this experiment. 

\begin{figure}[!t]
\begin{center}
\subfloat[]{\includegraphics[width = 2.2in]{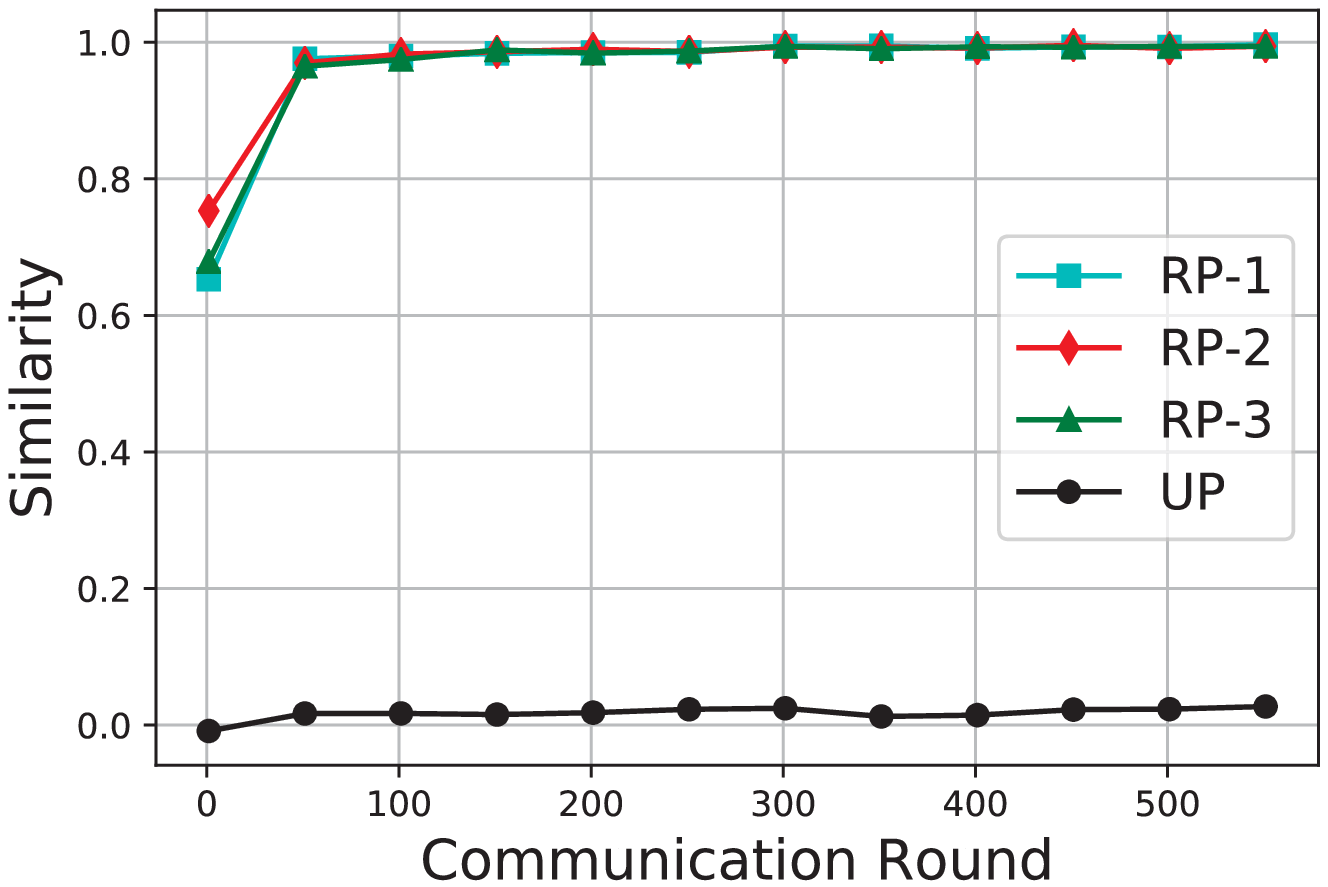}}
\subfloat[]{\includegraphics[width = 2.2in]{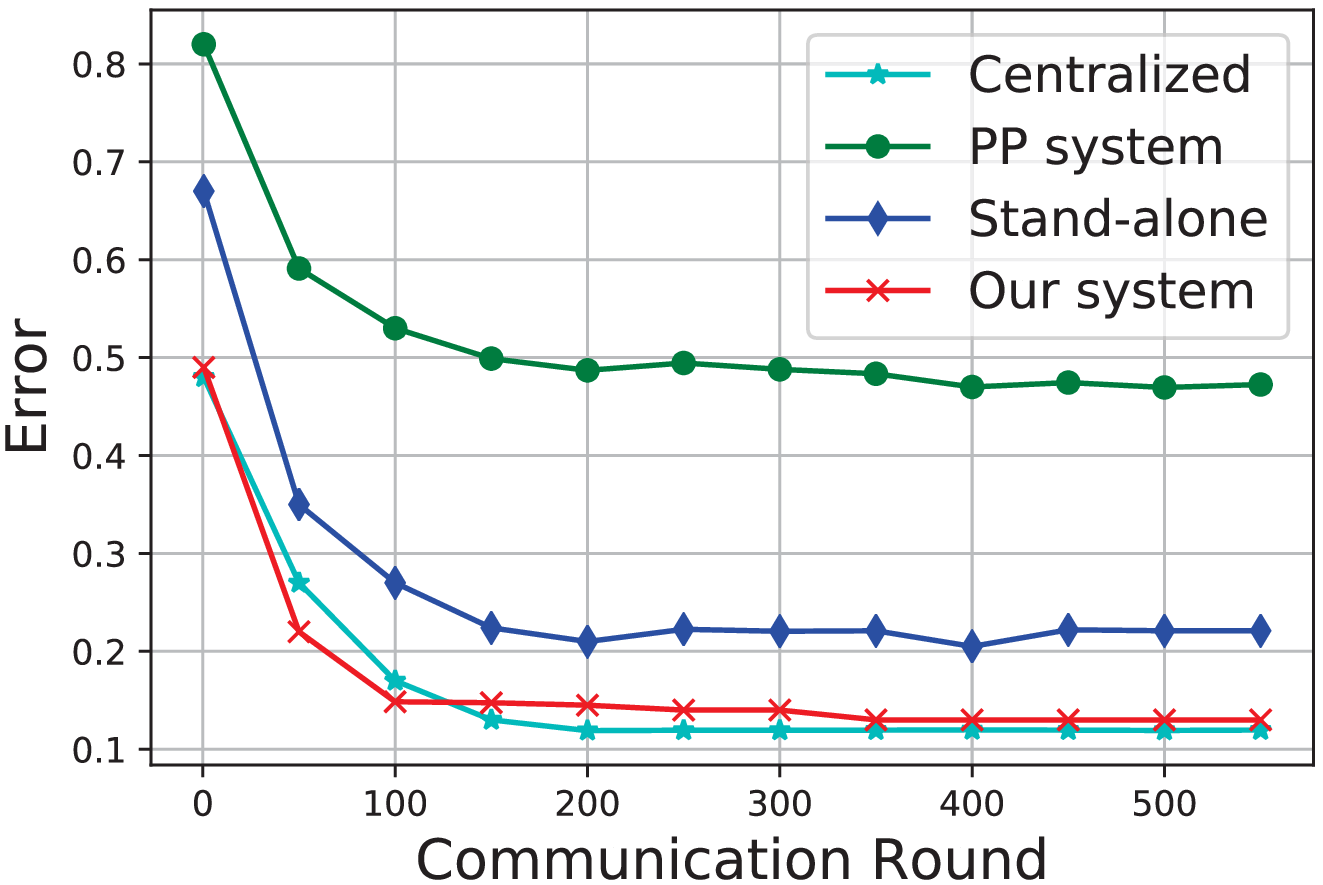}}
\subfloat[]{\includegraphics[width = 2.2in]{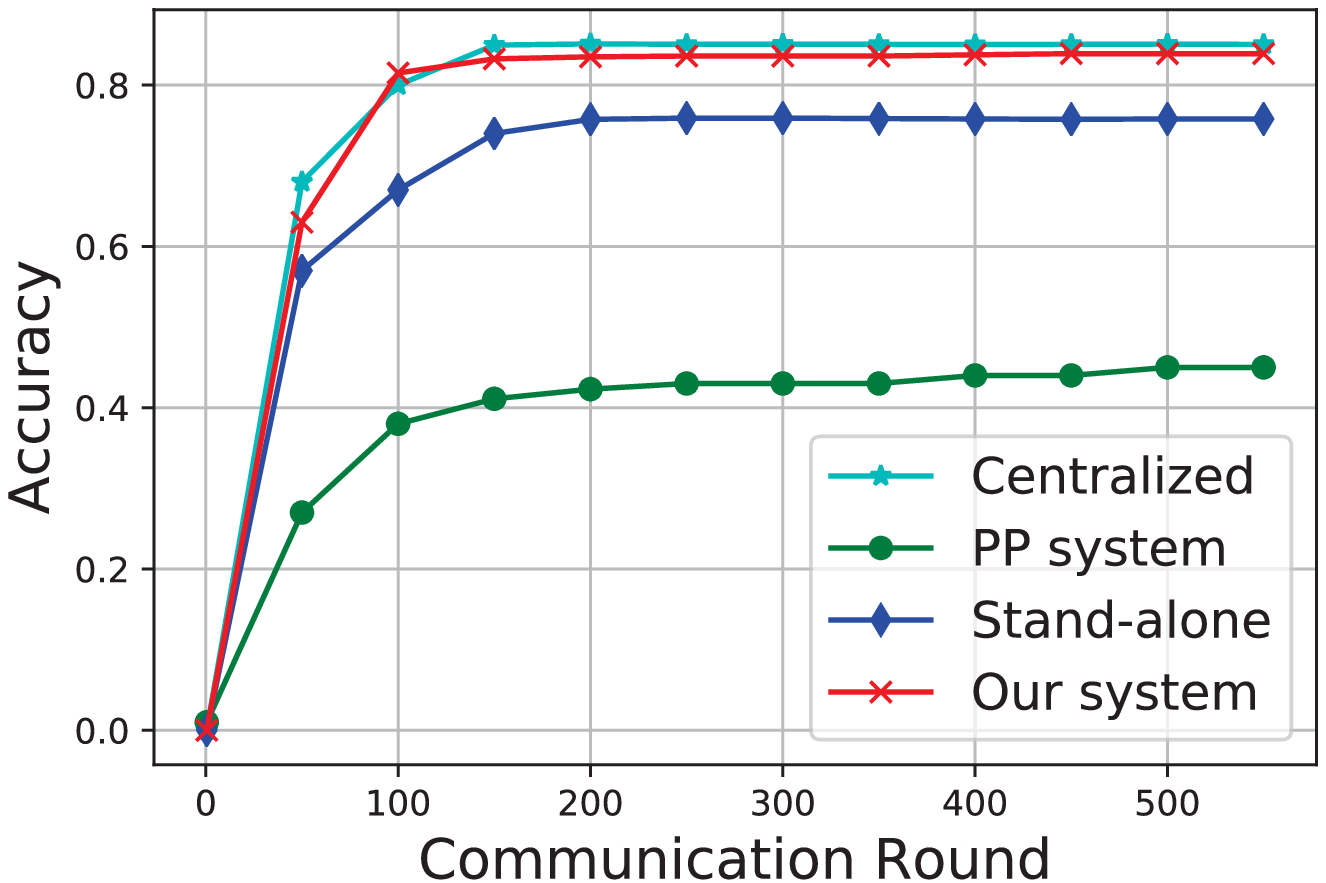}}\\
\caption{Results of logistic regression with CIFAR-10 dataset. (a) Weight similarities of the participants' weights to the model initiator's weight. (b) Test errors against communication round. (c) Accuracies against communication round.}
\label{fig:lr_cifar}
\end{center}
\end{figure}

\textit{Results:} The results of the above experimental settings are shown in Figure \ref{fig:lr_cifar}. Figure \ref{fig:lr_cifar}(a) depicts the weight similarity score between participants' and the model initiator's weight parameters against the communication rounds. As expected, the UP's and the model initiator's weight parameters are the least similar. The RPs' parameters are highly similar to the model initiator's parameters since their local models generate similar gradients that affect the weight updates similarly. Figure \ref{fig:lr_cifar}(b) and Figure \ref{fig:lr_cifar}(c) depict the training convergence of the model against the number of communication rounds. Our proposed system converges faster and achieves an accuracy closer to the centralized baseline. The stand-alone baseline has limited data and hence lower accuracy. The PP system does not filter out UPs during the computation of global parameters and as a result, it converges slowly and achieves the least accuracy.

\begin{figure}[!t]
\begin{center}
\subfloat[No. of local epochs = 5]{\includegraphics[width = 2.2in]{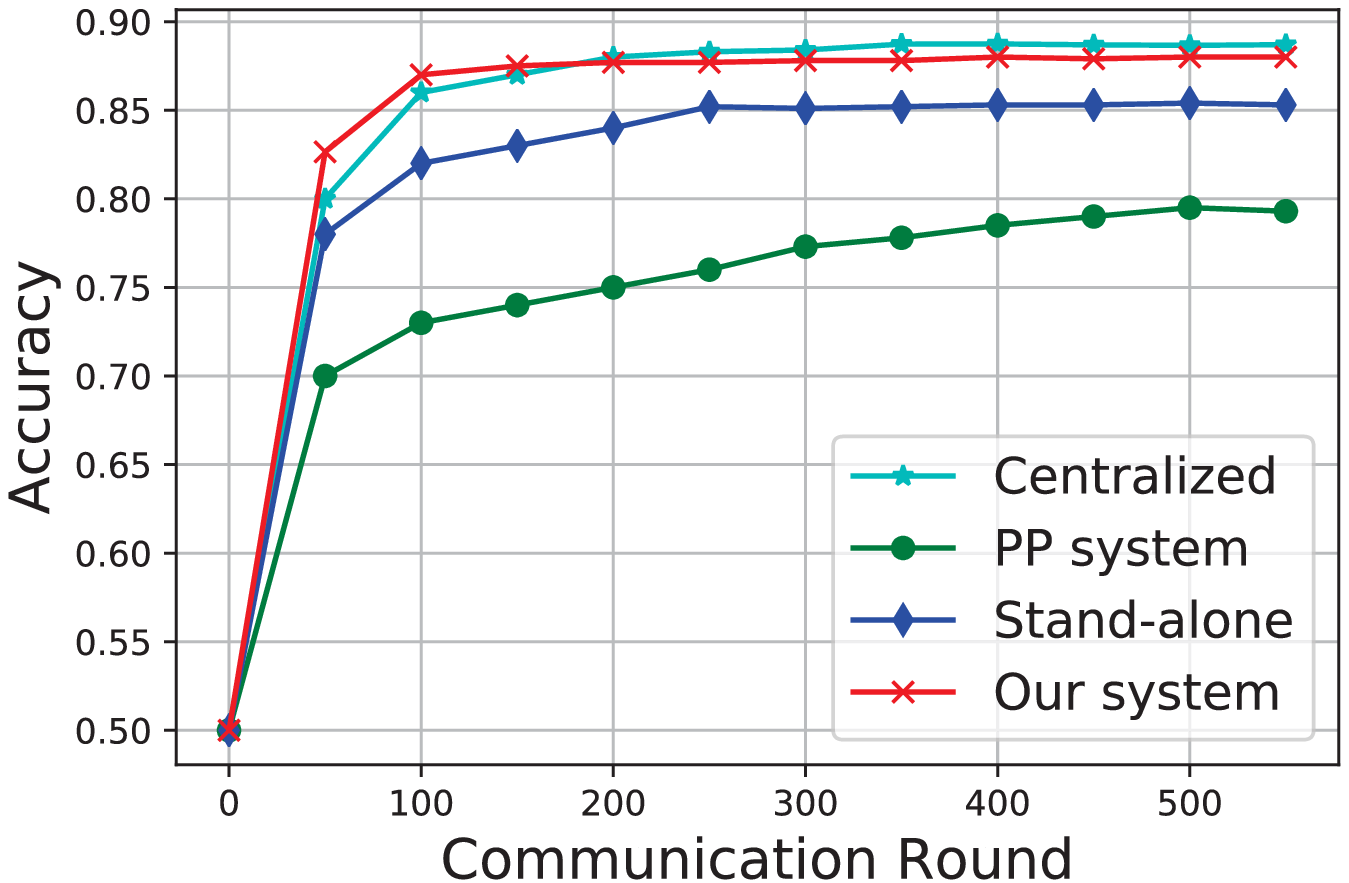}}
\subfloat[No. of local epochs = 10]{\includegraphics[width = 2.2in]{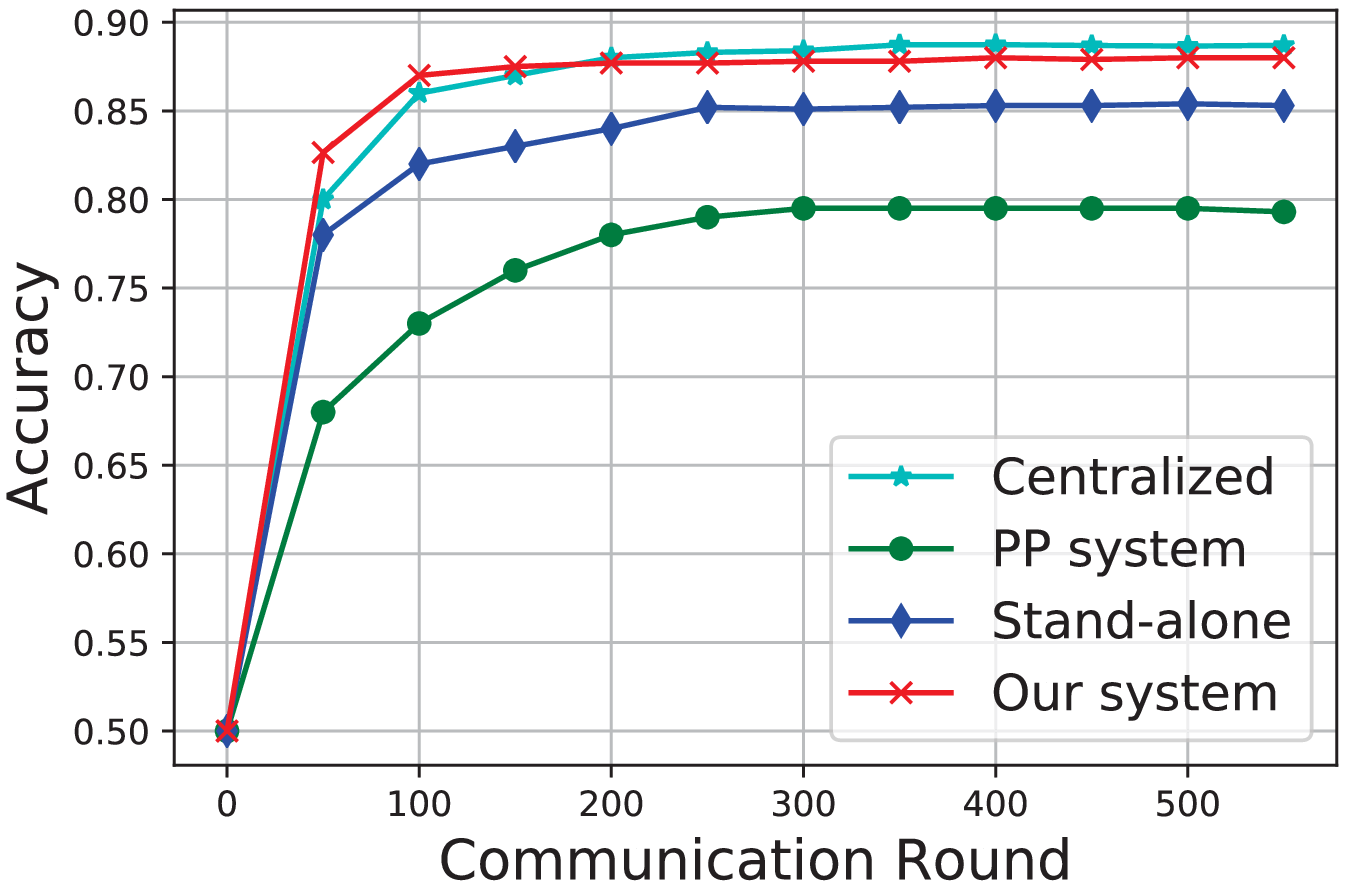}}
\subfloat[No. of local epochs = 50]{\includegraphics[width = 2.2in]{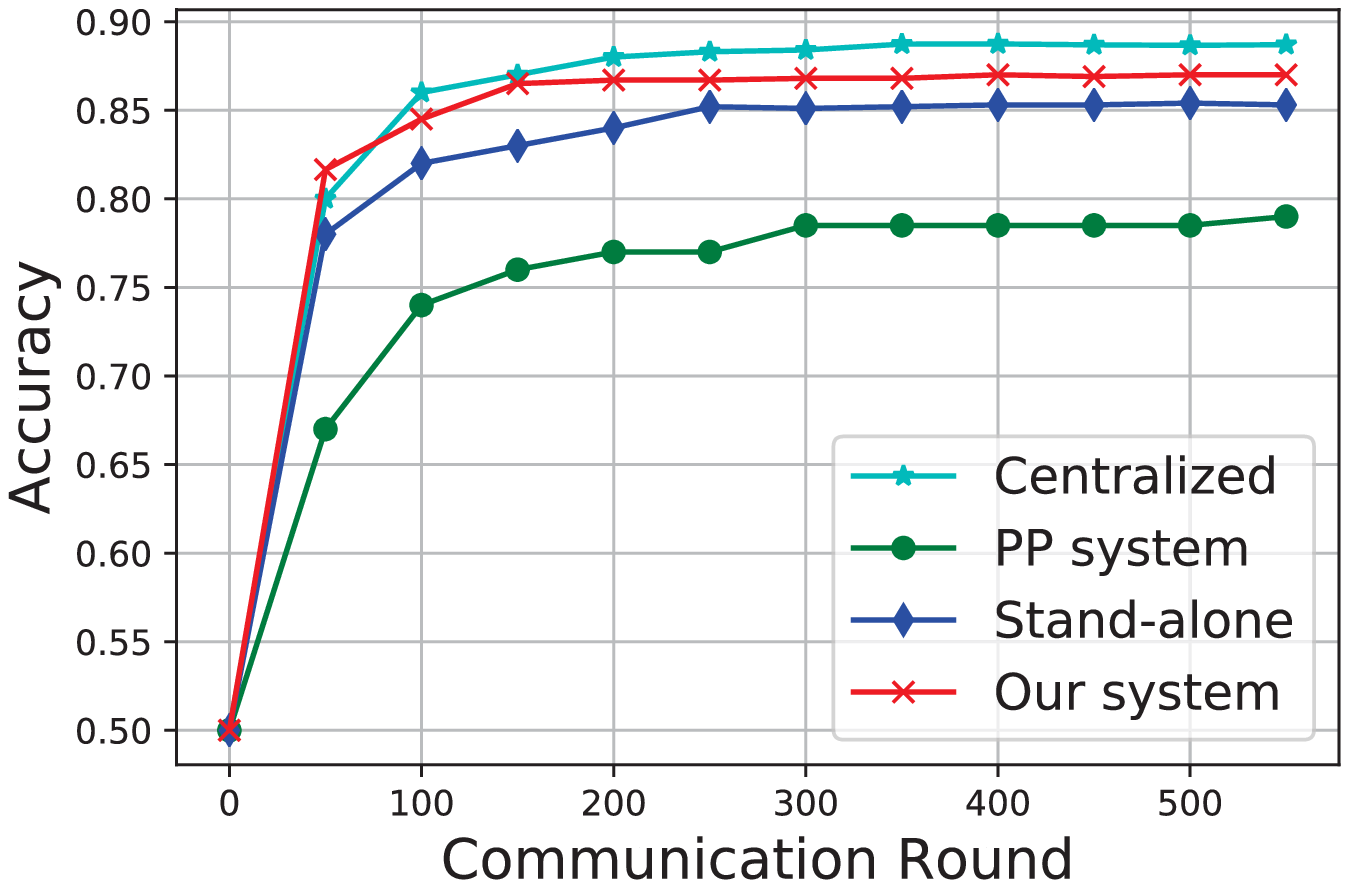}}\\
\caption{Results of MLP with CIFAR-10 dataset showing convergence against the number of communication rounds at different number of local epochs.}
\label{fig:nna_cifar}
\end{center}
\end{figure}

\subsubsection{Using an MLP}
We also evaluated our system using the CIFAR-10 dataset by implementing a two hidden layer MLP model each with 64 and 128 neurons, respectively. The implementation was done with Python and TensorFlow 2.1.0. The ReLU activation function was used for the hidden layers and the sigmoid for the output layer. In the experiment, the random seeds are set as follows: \textit{numpy.random.seed(15)}, \textit{random.seed(123)} and \textit{tensorflow.set\_random\_seed(12345)}. The SGD was used for learning with a batch size of 32 data items and a learning rate of $10^{-5}$. All the participants and the model initiator execute the same model on their local datasets and upload their encrypted intermediate results to the server with a varying number of local epochs. We dynamically varied the similarity score threshold in the range 0.05 to 0.95 in intervals of 0.15 after every 100 communication rounds during the global parameter updates by the server. 

\textit{Results:} The results of the MLP model with the CIFAR-10 dataset are presented in Figure \ref{fig:nna_cifar}. In Figure \ref{fig:nna_cifar}, we depict the training convergence of our system, PP system, and the baseline schemes against the communication rounds at different number of local epochs. In all the cases, our proposed system converges faster than all the other systems and attains an accuracy only bettered by the centralized baseline. However, at the local epoch of 50, the accuracy of our system slightly drops and as stated earlier, this could be due to improper mixing of parameters. The centralized baseline as expected achieves the best accuracy. The stand-alone baseline is less accurate as compared to ours and this is mainly because of the limited amount of the training data. The PP system does not filter out UPs and it indiscriminately includes parameters from all the participants during the global parameter updates which results in its slow convergence and the least accuracy.

\subsection{Results of Similarity Computation}
In Table \ref{tab:sim}, we present the execution time of the weight similarity score computation by the entities in our proposed scheme. We observe that the model initiator incurs the most computation overhead during the joint weight similarity score computation, which is mainly due to the encryption of its similarity computation component. The server simply performs a blinding operation and hence has the least computation overhead. The overhead for participants is associated with the multiplication operations they perform on the encrypted similarity computation components of the model initiator and the decryption of the final similarity score. Generally, the computation overhead is more for the neural network models as compared to the logistic regression models. This is because of the larger number of parameters associated with neural network models as compared to the logistic regression models.

\begin{table}
\caption {Run time for computation of weight similarity scores} \label{tab:sim}
\begin{center}
\begin{tabular}{|lcccc|}
    \hline
    
    \hline
    Dataset and Model & \multicolumn{2}{c}{MNIST} & \multicolumn{2}{c|}{CIFAR-10} \\
     & Logistic Regression & Neural Network & Logistic Regression & Neural Network \\    
    \hline
    Model Initiator(s) & 3.14 & 8.33 & 3.14 & 8.86 \\
    
    Participant(s) & 2.36 & 6.98 & 2.38 & 7.52 \\
    
    Server(s) & 0.19 & 0.63 & 0.19 & 0.65 \\
    \hline
    
    \hline
\end{tabular}
\end{center}
\end{table}

\section{Conclusion}\label{sec:conclusion}
In this work, we propose a multi-party privacy-preserving machine learning scheme that takes into account the data quality of the participants. The scheme utilizes the proposed weight similarity metric to filter out unreliable participants and integrates homomorphic encryption to prevent leakages to the server. In addition, participants upload their intermediate weights instead of gradients to prevent leakages to malicious participants. Therefore, our scheme is beneficial for privacy-preserving machine learning in environments where data quality matters. Possibilities for several future investigations are opened through this work. For instance, this scheme is designed to run synchronously, an asynchronous design can be a future possibility. Further investigation of the weight similarity under different untrainable parameters is another future possibility. Reducing the computation burden on the model initiator and the participants during weight similarity computation can be given more attention in the future. 
\bibliographystyle{elsarticle-num-names-alphsort}

\bibliography{mybibfile}


%
%
%

\end{document}